# Place Recognition Meet Multiple Modalities: A Comprehensive Review, Current Challenges and Future Development

Zhenyu Li[a,*], Tianyi Shang[b], Pengjie Xu[c], Zhaojun Deng[d]

[a]*The School of Mechanical Engineering, Qilu University of Technology (Shandong Academy of Sciences), Jinan 250300, China*
[b]*The School of Electronic Information Engineering, Fuzhou University, Fuzhou 350000, China*
[c]*The School of Mechanical Engineering, Shanghai Jiaotong University, Shanghai 201100, China*
[d]*College of Surveying and Geo-Informatics, Tongji University, Shanghai 200092, China*

---

## Abstract

Place recognition is a cornerstone of vehicle navigation and mapping, which is pivotal in enabling systems to determine whether a location has been previously visited. This capability is critical for tasks such as loop closure in Simultaneous Localization and Mapping (SLAM) and long-term navigation under varying environmental conditions. This survey comprehensively reviews recent advancements in place recognition, emphasizing three representative methodological paradigms: Convolutional Neural Network (CNN)-based approaches, Transformer-based frameworks, and cross-modal strategies. We begin by elucidating the significance of place recognition within the broader context of autonomous systems. Subsequently, we trace the evolution of CNN-based methods, highlighting their contributions to robust visual descriptor learning and scalability in large-scale environments. We then examine the emerging class of Transformer-based models, which leverage self-attention mechanisms to capture global dependencies and offer improved generalization across diverse scenes. Furthermore, we discuss cross-modal approaches that integrate heterogeneous data sources such as Lidar, vision, and text description, thereby enhancing re-silience to viewpoint, illumination, and seasonal variations. We also summarize standard datasets and evaluation metrics widely adopted in the literature. Finally, we identify current research challenges and outline prospective directions, including domain adaptation, real-time performance, and lifelong learning, to inspire future advancements in this domain. The unified framework of leading-edge place recognition methods, i.e., code library, and the results of their experimental evaluations are available at https://github.com/CV4RA/SOTA-Place-Recognitioner.

*Corresponding author

---

## 1. Introduction

Place recognition (PR) is a critical component of autonomous driving, enabling robust global localization, loop closure detection, and map consistency in GPS-denied environments [1], [2], [3]. By matching current sensory data with prior map information, it enhances localization accuracy, supports semantic understanding for decision-making, and improves resilience under appearance variations through multimodal integration [4], [5].

Place recognition has evolved as one of the most promising domains in artificial intelligence (AI), especially in autonomous driving [6], [7], [8], [5]. Place recognition, the task of identifying a previously visited location based on sensory input, constitutes a foundational capability in autonomous navigation [9], [10], SLAM (see Figure 1) [11], [12], [13], and lifelong vehicle operation. Accurate and robust place recognition enables loop closure detection, supports map consistency, and contributes significantly to the resilience of localization in dynamic and perceptually challenging environments. Over the past two decades, the field has undergone a substantial transformation, evolving from early geometry- and appearance-based methods to sophisticated deep learning architectures capable of capturing complex visual and multimodal correlations .

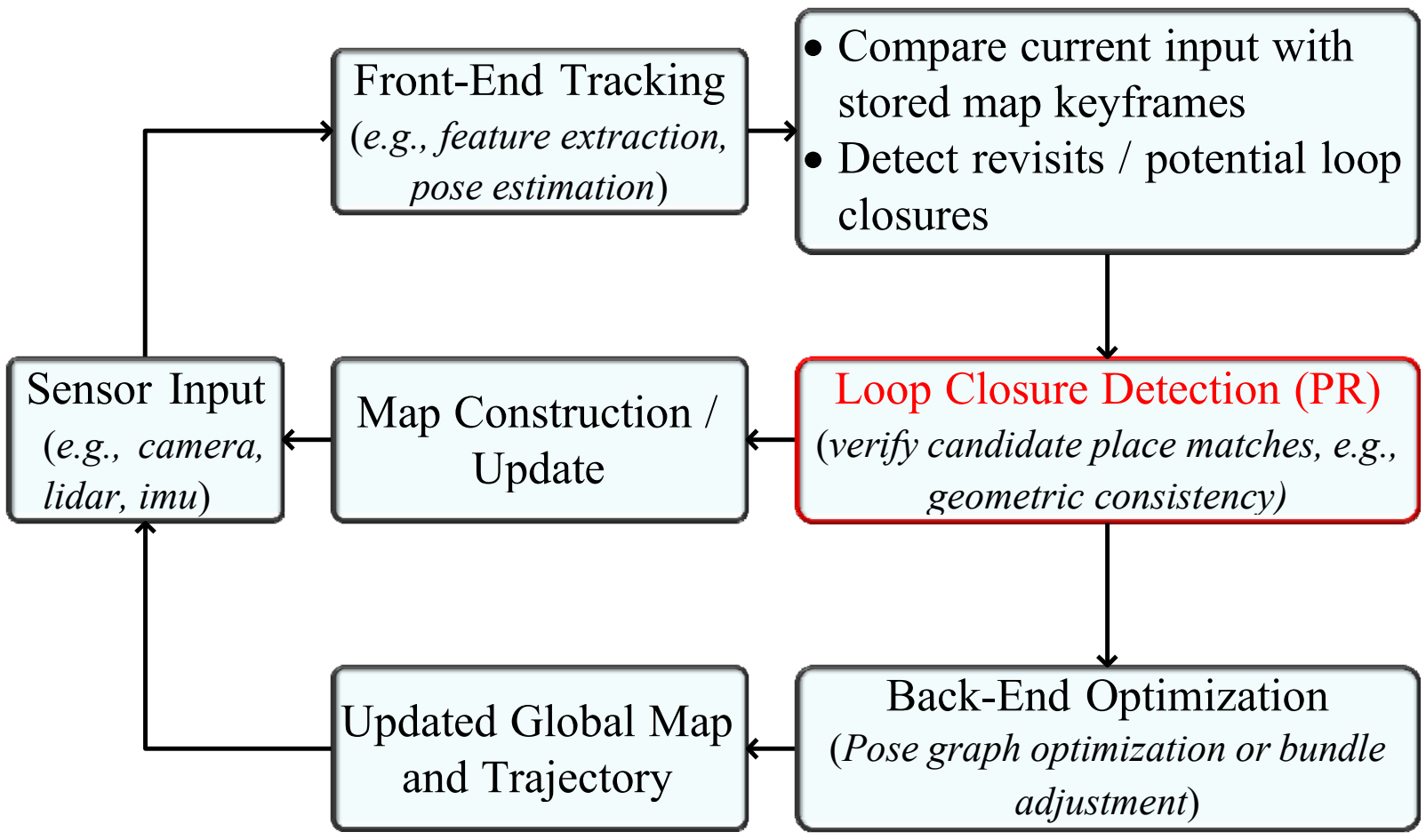

Figure 1: Place Recognition (PR) in SLAM. Place recognition plays a critical role in SLAM, particularly in detecting loop closures, which helps reduce accumulated drift by recognizing a previously mapped location, thus enabling the system to correct its pose estimate and improve map consistency.

Early approaches to place recognition predominantly relied on handcrafted features such as SIFT [14] and SURF [15], combined with Bag-of-Words (BoW) models and geometric verification [16]. While effective in controlled scenarios, these methods often suffer from limited robustness to large viewpoint or appearance changes. The emergence of Convolutional Neural Networks (CNNs) introduced a paradigm shift, offering superior feature representations and end-to-end learning capabilities. CNN-based methods significantly improved robustness and scalability, especially in large-scale urban environments and across diverse lighting or weather conditions [17].

In recent years, the field has seen another leap forward with the introduction of Transformer-based architectures [18],[19],[20]. Unlike CNNs, which capture local spatial patterns, Transformers employ self-attention mechanisms to model global contextual relationships, thereby enhancing generalization across varied environmental conditions. In parallel, the rise of cross-modal learning has further expanded the scope of place recognition, enabling systems to integrate visual data with other modalities such as Lidar, vision, or text description [21, 22, 23]. These advancements have markedly improved the robustness and versatility of place recognition systems, particularly under extreme environmental variability or sensor degradation.

Despite substantial progress, existing surveys often limit their scope to specific algorithmic families or modalities, leaving a fragmented understanding of the field. In contrast, this survey aims to provide a more holistic and systematic overview that spans CNN-based approaches, Transformer-based models, and cross-modal strategies. We examine each paradigm in depth, analyze their respective strengths and limitations, and discuss how they collectively contribute to the maturation of place recognition as a field. Furthermore, we consolidate information on benchmarks and evaluation protocols, and highlight open challenges and future directions that could shape the next generation of place recognition systems.

The contributions of this work lie fourfold:

- Compared to the previous survey of place recognition, such as [16], [17], [24], [25], [26], our overview is more comprehensive and in-depth. Specifically, our survey presents an overview of the primary methods from the early developments up to 2025, including visual, lidar, and cross-modal place recognition approaches. To my knowledge, this is the first methodological survey ever to include such a comprehensive review.

- This work provides a thorough introduction to place recognition, enumerates a variety of SOTA approaches, and emphasizes the discussion regarding the application from the perspective of autonomous driving vehicles.

- In this work, most publicly accessible place recognition methods are merged into a single code base for the first time. In addition, experimental results are provided for convenient reference.

- This survey discusses all the solutions, summarizes the advantages and shortcomings of the existing methods, and looks forward to the future development direction.

To the best of our knowledge, this survey represents the first comprehensive survey of place recognition methods, including VPR, LPR, and CMPR methods, developed over the past decade. The remainder of the survey is organized as follows: Section 2 discusses the practical development and technical challenges associated with place recognition. Section 3 provides a summary of place recognition methods from the past two decades, including visual place recognition (VPR), LiDAR place recognition (LPR), and cross-modal place recognition (CMPR) schemes. Section 4 introduces existing datasets, evaluation metrics, and the assessment of current place recognition methods. Section 5 evaluates the results of VPR, LPR, and CMPR. Section 6 provides challenges and solutions for each modality. Finally, Section 7 concludes the paper. The structural framework of this survey is illustrated in Figure 2, and the method's evolution over time is depicted in Figure 3.

## 2. Background

### 2.1. Development

The origins of SLAM date back to 1986, when Smith and Cheeseman formally introduced the problem as a means for robots to build maps of unknown environments while localizing themselves. The 1990s saw the popularization of EKF-SLAM, which used Extended Kalman Filters for probabilistic mapping [27]. However, due to its computational limitations in large-scale settings, the 2000s introduced GraphSLAM [28] and FastSLAM [29], both more scalable and efficient. In 2004, the release of GMapping [30] made real-time 2D laser-based mapping widely accessible.

In 2011, the rise of Visual SLAM (vSLAM) marked a major shift. ORB-SLAM [31], introduced in 2015, became a benchmark for feature-based monocular SLAM. Simultaneously, Direct Sparse Odometry (DSO) [32], introduced in 2016, showcased the viability of direct methods using pixel intensities. More recently, from 2018 onward, deep learning methods like DeepVO [33], VLocNet++ [34], and Transformer-based SLAM have emerged [35], enabling semantic un-derstanding and multi-sensor fusion (Lidar, IMU, vision) .

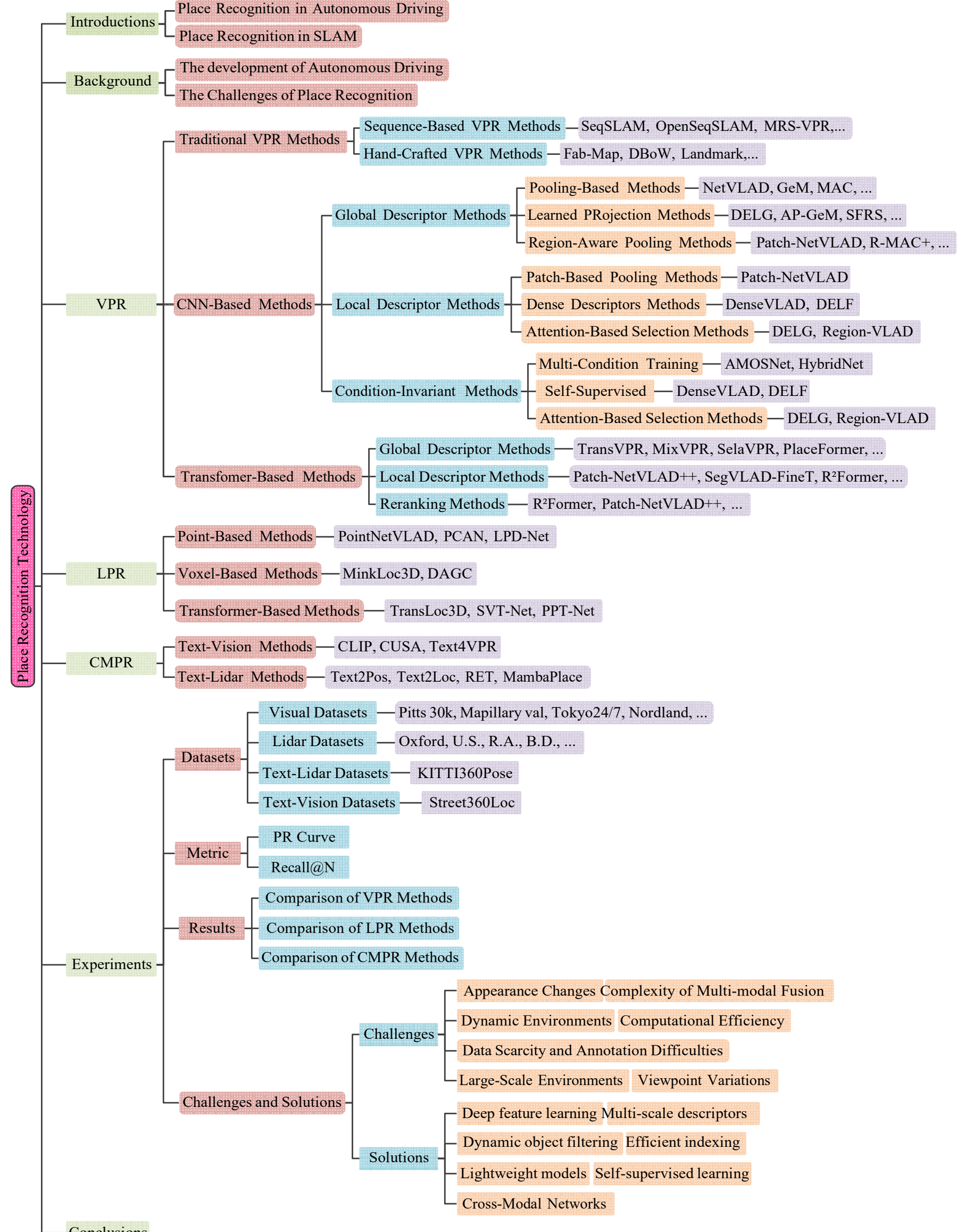

Figure 2: Structural framework of this survey.

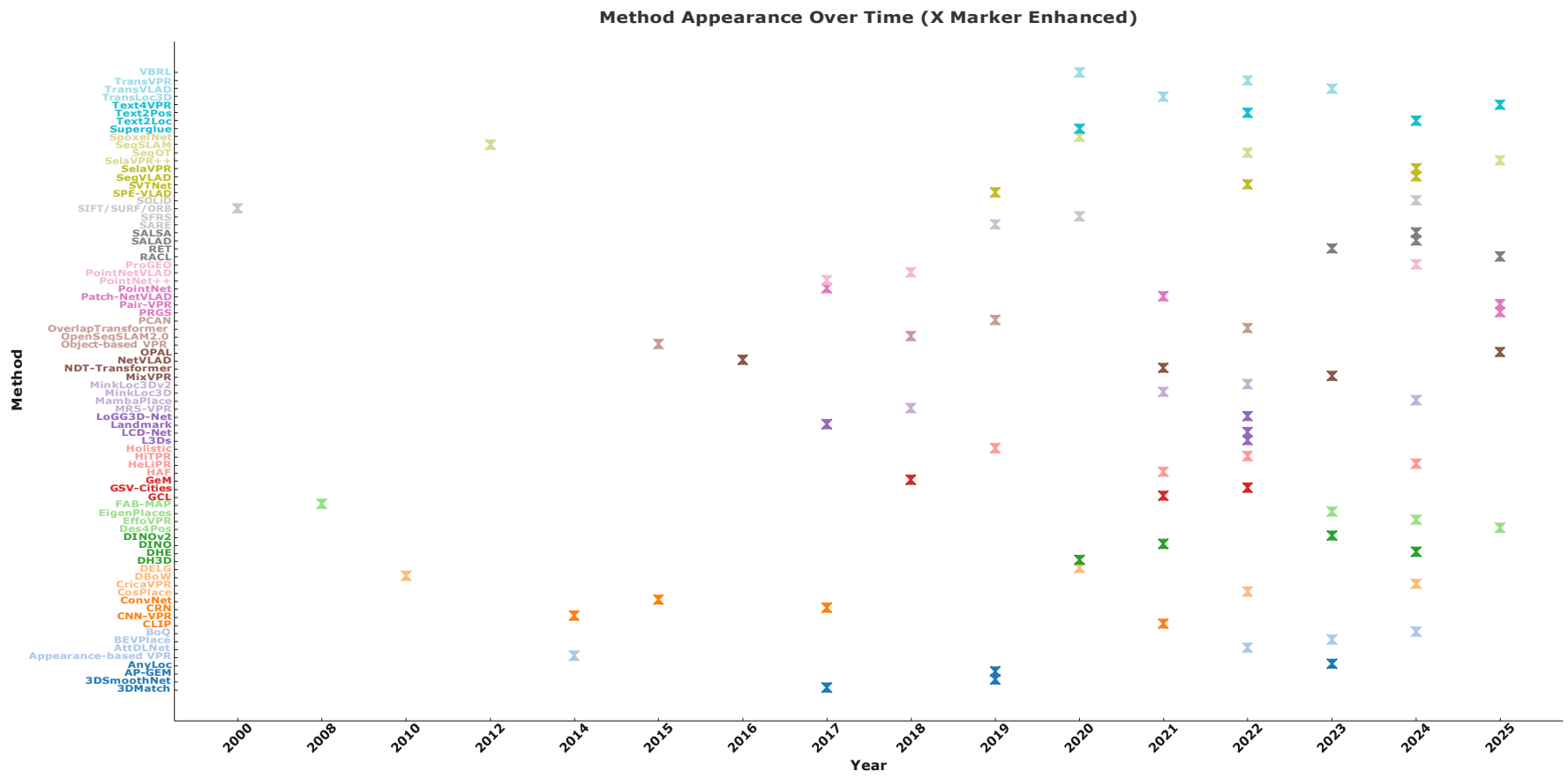

Figure 3: Evolution of VPR, LPR, and CMPR methods over time.

Meanwhile, autonomous vehicles began their ascent in the 1980s, with early prototypes like Navlab (Carnegie Mellon) and Mercedes-Benz's Prometheus Project. A breakthrough came in 2004 with DARPA's Grand Challenge, followed by the Urban Challenge (2007). These events catalyzed the birth of Google's self-driving car project in 2009, later known as Waymo.

China entered the race during the 2010s: Baidu Apollo launched in 2017, offering an open-source autonomous driving platform. Huawei introduced ADS 2.0 in 2023, promoting map-free driving. Xpeng (NGP) and NIO (NOP+), since 2021, have developed high-level assisted driving on highways and urban roads, often leveraging SLAM techniques for positioning and scene understanding.

By 2024, SLAM and autonomous driving have become inseparable: SLAM powers real-time localization, loop closure, map maintenance, and route relocalization, serving as the perception backbone for increasingly capable autonomous systems.

The timeline of autonomous vehicle and SLAM development is shown in Figure 4 and Figure 5.

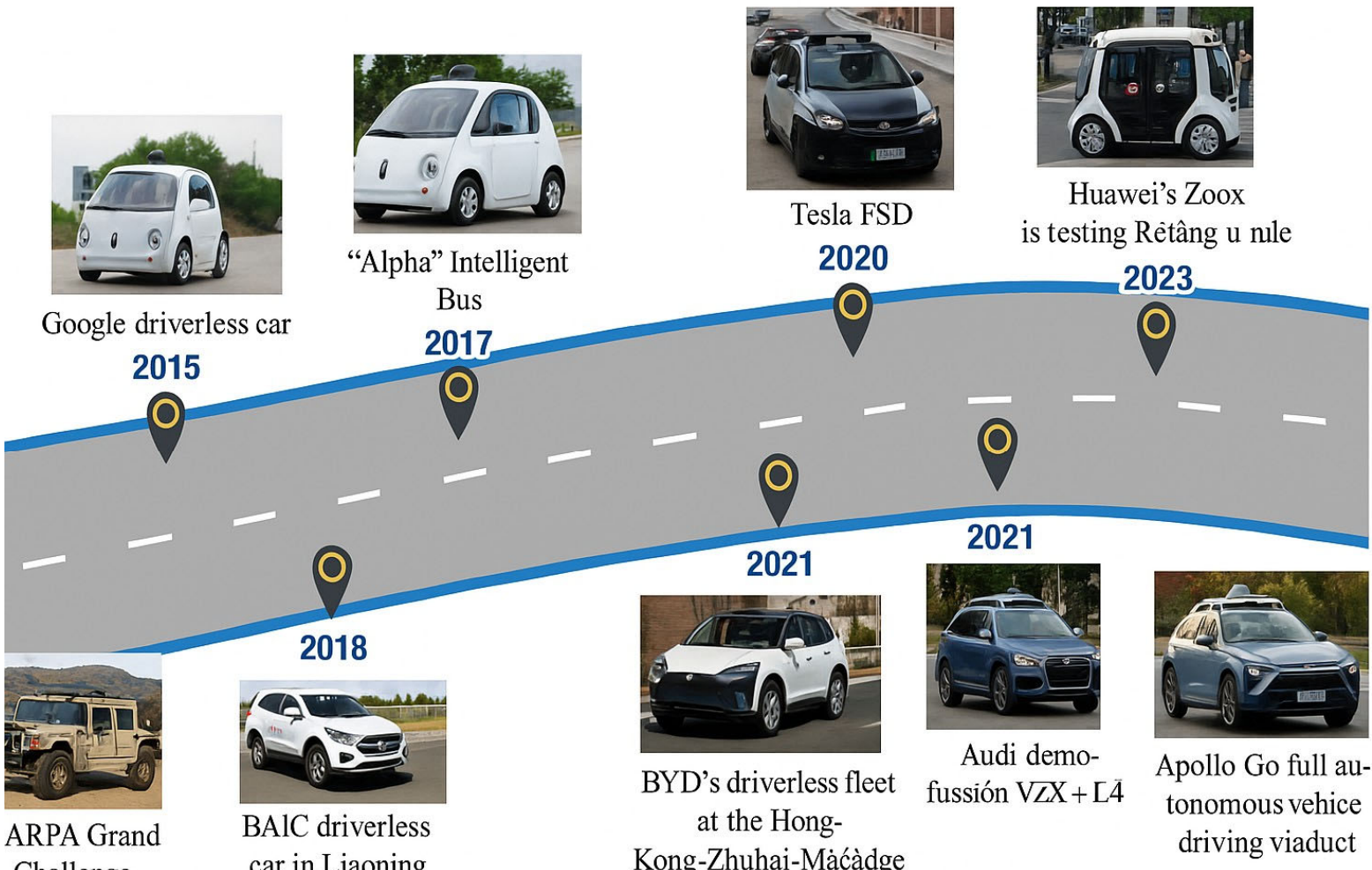

Figure 4: The development timeline of autonomous driving application technology.

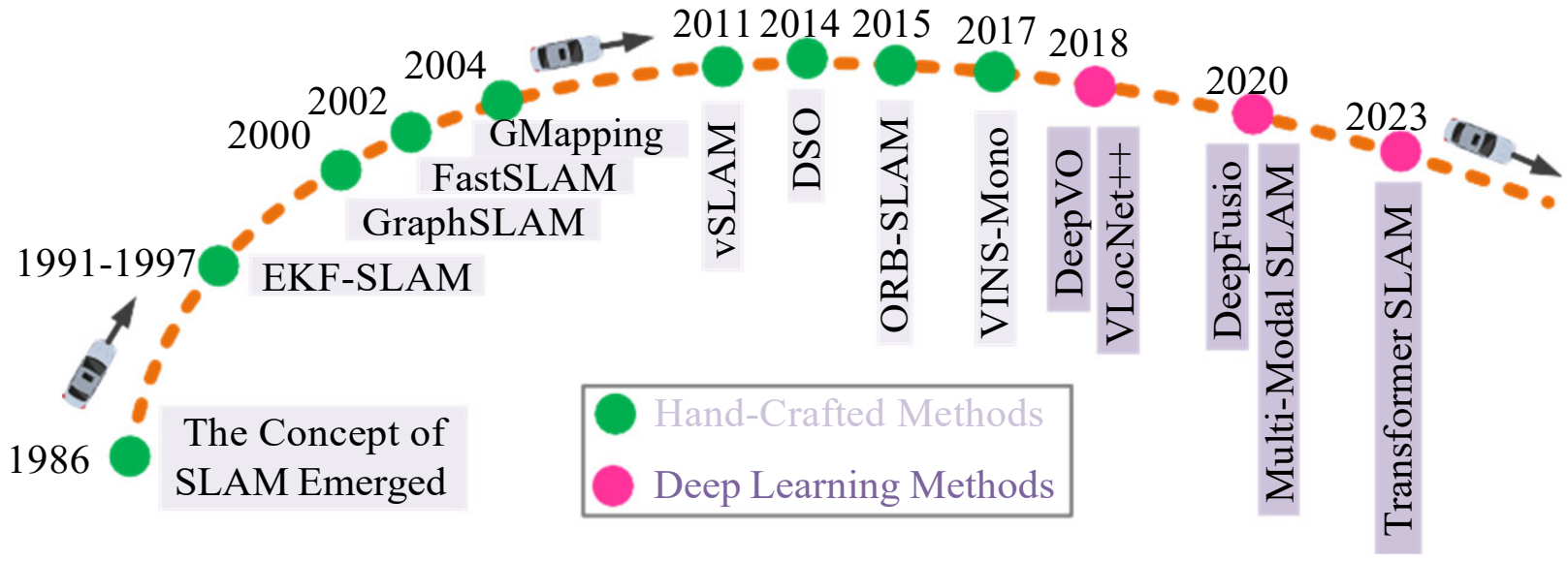

Figure 5: The mainstream SLAM technologies.

### *2.2. Significance and challenges*

As autonomous driving technology advances from controlled prototypes to real-world deployment, precise environmental awareness becomes increasingly critical. One of the core components that facilitates this awareness is place recognition—the ability of a vehicle to determine whether it has previously visited a specific location. Although this function may seem conceptually straightforward, it underpins many essential capabilities required for autonomous driving to be safe, reliable, and efficient. This article offers a comprehensive discussion on the significance of place recognition within autonomous driving systems, examining its impact on localization, mapping, planning, safety, scalability, and user experience.

In numerous real-world driving scenarios, such as urban canyons, tunnels, underground parking facilities, or areas with limited satellite visibility, GPS signals can be degraded or lost. Relying exclusively on GNSS (Global Navigation Satellite Systems) for positioning in these conditions is impractical. Place recognition allows vehicles to determine their location by matching current sensor data (e.g., visual images or LiDAR scans) with a previously recorded map or database of known locations. This approach offers an independent, infrastructure-free method for accurately determining location, serving as a reliable fallback or complement to GPS and dead reckoning techniques. Moreover, even in environments with strong GPS reception, place recognition safeguards against accumulated drift from visual-inertial odometry or wheel encoders, ensuring that the vehicle's internal localization remains aligned with the real world.

SLAM is a fundamental capability for autonomous vehicles, especially in uncharted or dynamic environments [36], [37],[38], [39]. Within SLAM systems, loop closure detection—identifying when a vehicle returns to a previously visited location—is essential for correcting trajectory drift and maintaining a coherent global map. Place recognition functions as the mechanism for detecting loop closures. When a match is identified between the current scene and a stored map location, the system can retroactively correct errors in the vehicle's estimated path and update the map accordingly. This process prevents long-term drift and ensures the geometric consistency of the map, which is essential for navigation and long-term autonomy.

Summarizing the above development achievements, we can summarize the challenging problems encountered by autonomous driving and SLAM as follows:

- Robustness to Appearance Changes: Autonomous driving systems rely on place recognition, but environmental changes like lighting and weather cause significant variations in appearance.
- Viewpoint and Perspective Variation: Recognizing places from different viewpoints and scales is difficult in SLAM, as locations can appear vastly different from various angles.
- Dynamic and Repetitive Environments: False loop closures occur when locations appear visually similar in urban or repetitive environments.
- Real-Time Constraints in Large-Scale Systems: SLAM systems must operate in real-time in large environments, demanding efficient place recognition techniques.
- Sensor Modality and Cross-Modal Matching: Integrating multiple sensors like vision, LiDAR, and radar improves SLAM robustness but creates challenges in aligning data from different modalities.

Various factors affecting lane visibility in the real-world environment, as Fig. 6:

## 3. Place Recognition Technology

The development of place recognition technology plays a crucial role in implementing autonomous and assisted driving systems. The general structure of place recognition technology can be illustrated by Figure 7. Researchers are diligently working to enhance the performance of place recognition. Current methods for place recognition can be categorized into visual place recognition, Lidar place recognition, and cross-modal place recognition.

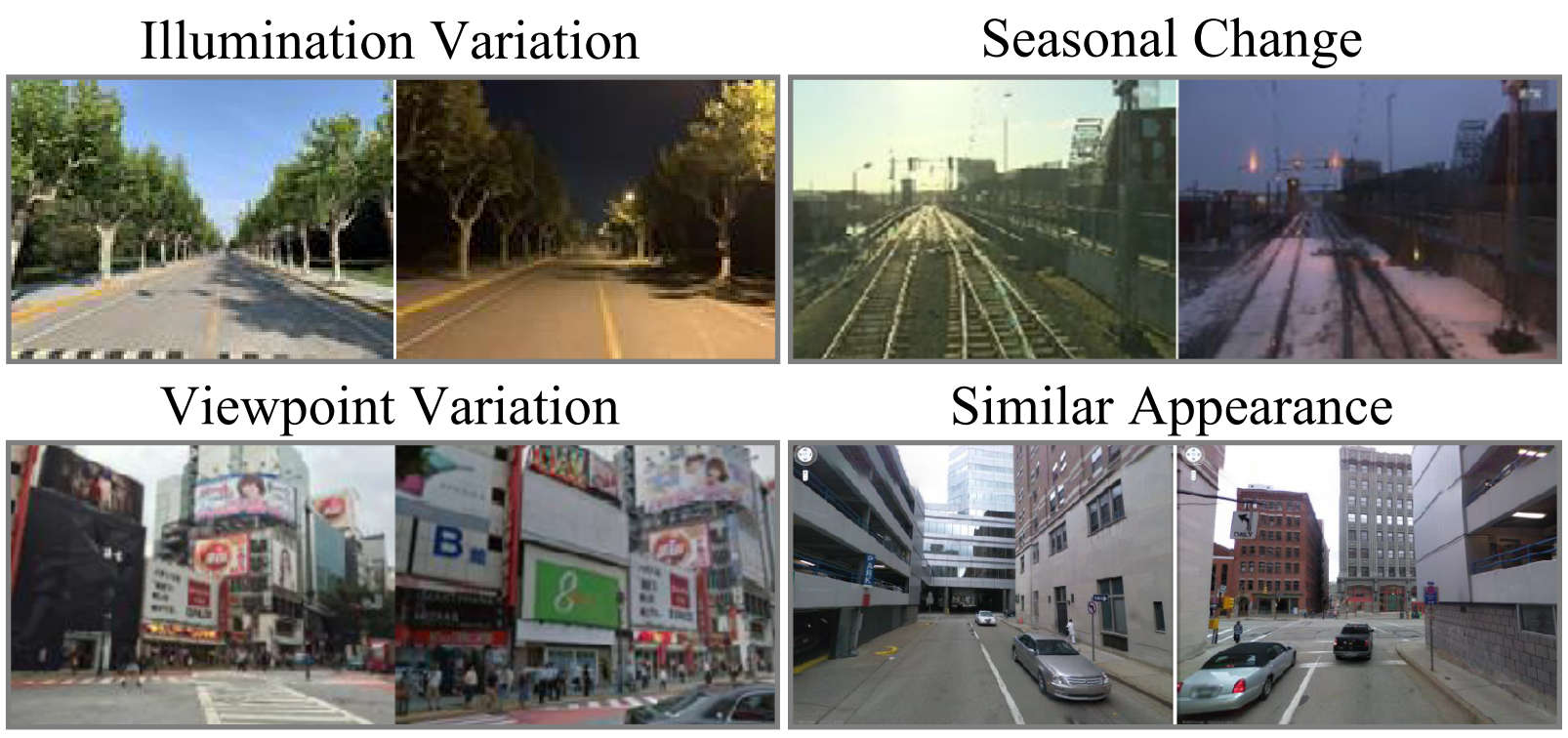

Figure 6: Various factors affecting lane visibility in the real-world environments.

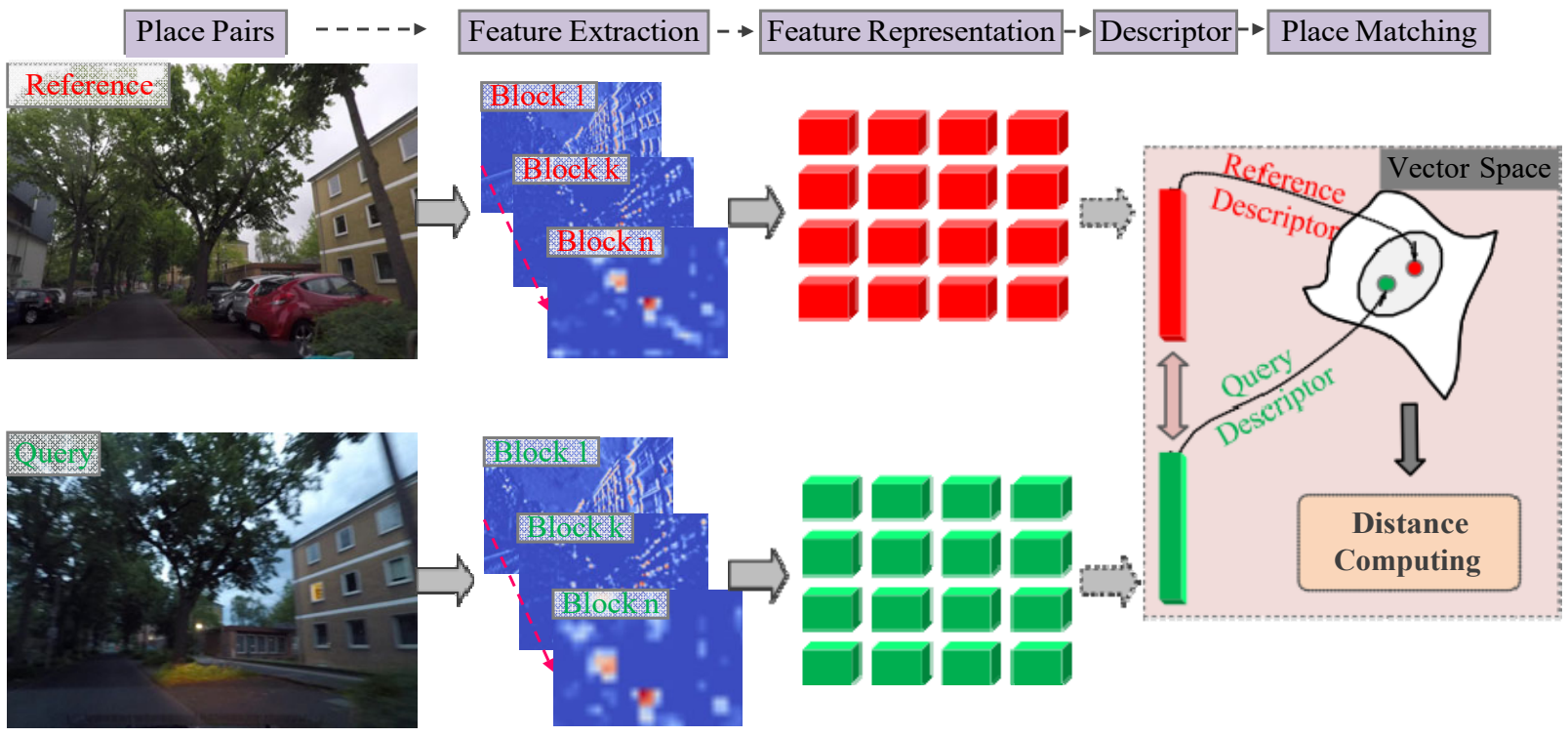

Figure 7: The general structure of place recognition technology.

### 3.1. *Visual Place Recognition (VPR)*

VPR refers to recognizing a specific location or place in a visual scene (usually an image or video) based on a prior experience or dataset of known places. As shown in Figure 8, the evolution of VPR methods is demonstrated.

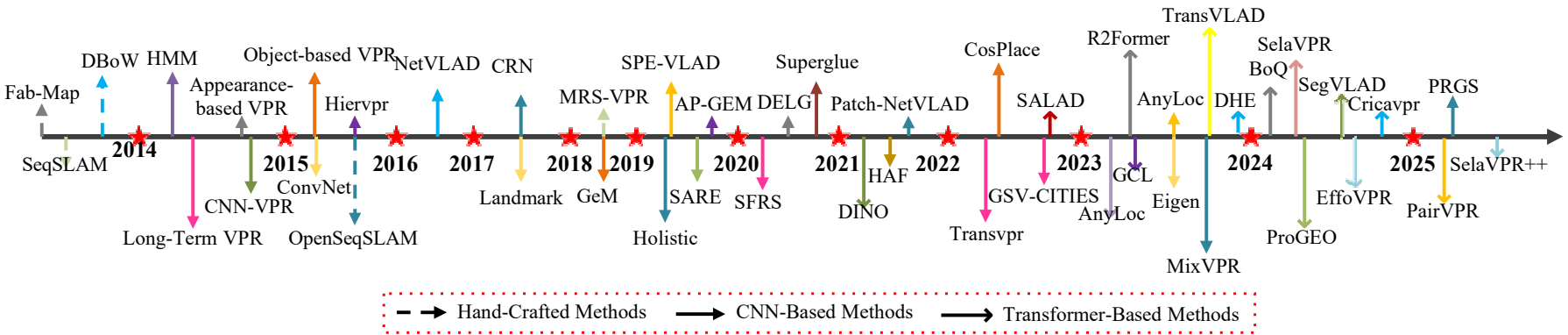

Figure 8: The different mainstream methods in the field of VPR in the recent decade. According to the feature encoding method, VPR methods are subdivided into manual engineering, CNN-based, and Transformer-based methods.

#### 3.1.1. *Sequence-Based VPR Methods*

Sequence-Based Visual Place Recognition (SeqVPR) methods depend on the comparison of image sequences rather than individual images to identify specific locations. These methods leverage temporal and sequential information to enhance robustness against changes in environmental conditions, such as variations in lighting, changes in viewpoint, and the presence of dynamic elements. Sequence-based approaches have proven particularly effective for long-term and large-scale localization tasks, where recognition based on individual images may falter due to these variations.

SeqSLAM [40] introduces a sequence-matching approach in which place recognition is based on the temporal alignment of sequential image frames. This method effectively addresses long-term environmental changes, such as seasonal variations and time-of-day differences, making it well-suited for mobile robots and autonomous systems that navigate over extended periods. OpenSeqSLAM2.0 [41] is an open-source implementation of the SeqSLAM algorithm, providing greater accessibility and flexibility for researchers. It supports further experimentation and customization, which is valuable for various place recognition tasks in both structured and dynamic environments. MRS-VPR [42] extends the SeqSLAM framework by incorporating multi-resolution image sequences. This method enhances the ability to match places across different spatial scales and provides improved robustness when dealing with changing environments.

### *3.1.2. Hand-Crafted Feature Methods*

Handcrafted feature engineering methods in Visual Place Recognition (HfeVPR) involve the extraction of distinctive visual features from images, which are then used for matching and identifying places. These methods are based on traditional computer vision algorithms, such as SIFT, SURF, and ORB, which are designed to detect and describe keypoints or local features that are invariant to scale, rotation, and partial occlusion.

FAB-MAP [43] utilizes a probabilistic approach based on the Bag-of-Words (BoW) model, where local features, such as SIFT, are clustered to form a visual vocabulary. DBoW [44] is an enhancement of the BoW model, utilizing binary feature descriptors (such as ORB) to represent images as visual words. This approach is computationally efficient and suitable for real-time applications.

Table 1: Summary of traditional non-deep learning VPR methods.

| Method | Type | Jour./Conf. | Key Features | Strengths | Year |
|---|---|---|---|---|---|
| SeqSLAM [40] | SeqVPR | ICRA | Sequential matching, temporal information | Long-term robustness | 2012 |
| FAB-MAP [43] | HfeVPR | IJRR | Bag-of-Words, probabilistic mapping | Probabilistic mapping | 2008 |
| DBoW [44] | HfeVPR | TRO | Binary descriptors, fast retrieval | Efficient performance | 2010 |
| SIFT/SURF/ORB | HfeVPR | – | Local feature, keypoint matching | Scale, rotation, and occlusion robustness | 2000s-2010s |
| MRS-VPR [42] | SeqVPR | ICRA | Multi-resolution matching | Multi-resolution adaptability | 2018 |
| OpenSeqSLAM2.0 [41] | SeqVPR | IROS | Sequential matching | Consistent context | 2018 |

**Remark 1:** *Sequence-based and handcrafted feature methods struggle with dynamic environments and extreme changes in conditions, as they rely on predefined rules and local features. Unlike deep learning methods, they lack the adaptability, robustness, and scalability needed to handle large, diverse datasets and complex environments.* The summary of traditional non-deep learning VPR methods is shown in Table 1.

### *3.1.3. CNN-Based VPR Methods*

CNN-based methods in VPR leverage CNNs to automatically learn hierarchical features from raw image data, eliminating the need for manual feature extraction. These methods excel at handling complex environmental changes, such as variations in lighting, viewpoint, and occlusions, by learning robust and discriminative representations. According to different feature extraction methods, CNN-based VPR can be divided into global descriptor method, local descriptor method, and conditional invariance method. The global/local feature-based VPR method can be described as shown in Figure 9. The detail category is shown in Table 2.

**A. Global Descriptor Methods:** Global descriptor methods focus on generating a single, fixed-size feature vector that represents the entire image.

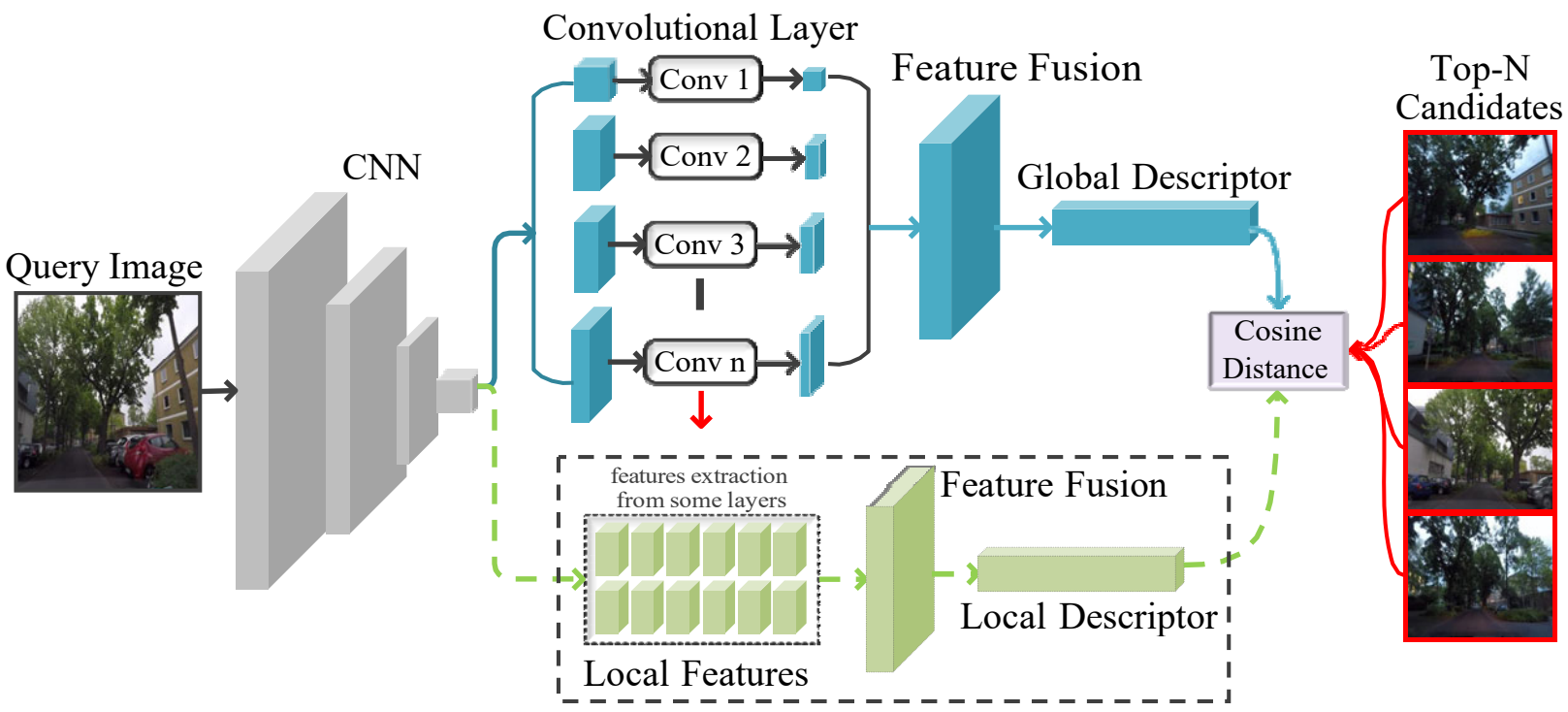

Figure 9: CNN-based VPR paradigm. The figure describes two different feature representation methods: the method based on local features and the method based on global features.

Appearance-based VPR [47] primarily focuses on unsupervised learning techniques to allow systems to recognize places reliably, even when environmental conditions (e.g., lighting, weather, time of day) change. NetVLAD [49] combines CNNs with the Vector of Locally Aggregated Descriptors (VLAD) method. It aggregates local features extracted by CNNs into a global descriptor that is robust to viewpoint and environmental changes. GeM [52] proposes a fully automated pipeline to fine-tune CNNs for image retrieval and place recognition using no manual annotations. It introduces a robust training framework based on mining hard positive and negative image pairs from large-scale, unlabeled photo collections. SPE-VLAD [53] integrates a spatial pyramid structure into the NetVLAD framework to capture multi-scale spatial information. By partitioning images into multiple scales and aggregating features at each level, SPE-NetVLAD captures both local and global contextual information, enhancing robustness to viewpoint and appearance changes. DELG [55] simultaneously extracts global and local features from images using a unified CNN architecture. This integration aims to combine the advantages of both feature types for improved image retrieval performance. Holistic [56] proposes a VPR system utilizing a lightweight CNN architecture, specifically AlexNet trained on the Places365 dataset. This choice balances performance and computational efficiency, making it suitable for deployment on mobile robots and embedded systems. MixVPR [58] utilizes feature maps from pre-trained backbones as global features, and incorporates a global relationship between elements in each feature map through a cascade of feature mixing, eliminating the need for local or pyramidal aggregation techniques. EigenPlaces [59] addresses the challenge of viewpoint shifts by clustering training data into classes where each class contains images depicting the same scene from different viewpoints. This approach ensures that the model learns global descriptors that are invariant to perspective changes, improving its ability to recognize places from various angles. CosPlace [60] addresses the challenges of applying visual geo-localization (VG) techniques to large-scale urban environments. The authors introduce a new dataset, San Francisco eXtra Large (SF-XL), which is 30 times larger than previous datasets. GSV-CITIES [61] provides a significant advancement in the field of visual place recognition by offering a comprehensive dataset and a novel aggregation technique, facilitating more accurate and efficient localization in urban environments.

Table 2: CNN-based VPR Methods.

| Method | Published In | Key Features | Type |
|---|---|---|---|
| CNN-VPR [45] | ArXiv 2014 | CNN for feature extraction and image matching | Local-based |
| ConvNet [46] | IROS 2015 | Hierarchical CNN representations for place recognition | Local-based |
| Appearance-based VPR [47] | ACRA 2014 | Unsupervised features for condition-robust recognition | Global-based |
| Object-based VPR [48] | ICRA 2015 | RGB-D object-level matching | Local-based |
| NetVLAD [49] | CVPR 2016 | VLAD with CNN features for global descriptors | Global-based |
| CRN [50] | CVPR 2017 | Context-aware local feature reweighting | Local-based |
| Landmark [51] | ACCV 2017 | Spatial landmark distribution representation | Local-based |
| GeM [52] | TPAMI 2018 | Learnable pooling for global representation | Global-based |
| SPE-VLAD [53] | TNNLS 2019 | Spatial pyramid-enhanced NetVLAD | Global-based |
| SFRS [54] | ECCV 2020 | Region similarity estimation for fine-grained retrieval | Local-based |
| DELG [55] | ECCV 2020 | Unified extraction of local and global features | Global-based |
| Holistic [56] | TRO 2019 | Lightweight AlexNet for embedded VPR | Global-based |
| Patch-NetVLAD [57] | CVPR 2021 | Patch-level aggregation from NetVLAD residuals | Local-based |
| MixVPR [58] | WACV 2023 | Feature mixing for compact global descriptors | Global-based |
| EigenPlaces [59] | ICCV 2023 | Viewpoint-invariant global descriptor learning | Global-based |
| CosPlace [60] | CVPR 2022 | Classification-based training with city-scale dataset | Global-based |
| GSV-Cities [61] | Neuro 2022 | Urban-scale dataset with Conv-AP aggregation | Global-based |

**B. Local Descriptor Methods:** Local descriptor methods in VPR focus on extracting and matching local features from images to identify places. These methods often emphasize robustness to local environmental variations, such as changes in viewpoint, lighting, and occlusion.

CNN-VPR [45] leverages the powerful feature extraction capabilities of CNNs to automatically identify and match locations in images. ConvNet [46] offers a significant improvement over traditional hand-crafted features in the domain of place recognition. Their ability to learn robust, hierarchical representations directly from data makes them particularly suited for dynamic and challenging real-world environments. Object-based VPR [48] improves place recognition using RGB-D maps (color and depth maps) by integrating object-based recognition techniques. The goal is to leverage both the visual and depth information to identify specific places in an environment, focusing on objects within the scene as key features for recognition. CRN [50] dynamically adjusts the importance of local features based on their surrounding context, thereby improving the discriminative power of image representations for geo-localization tasks. by incorporating spatial distribution information of landmarks within images, Landmark [51] addresses the limitations of traditional CNN-based global descriptors, which often overlook the spatial arrangement of features, leading to reduced robustness under varying viewpoints and environmental conditions.

SFRS [54] estimates and refines image-to-region similarities without manual annotations. By decomposing images into multiple sub-regions (e.g., halves and quarters), the model learns to associate specific regions of query images with corresponding regions in reference images, enhancing the granularity of feature learning. Patch-NetVLAD [57]derives patch-level features from NetVLAD residuals, enabling the aggregation and matching of deep-learned local features over the feature-space grid. This contrasts with traditional local keypoint features that rely on fixed spatial neighborhoods.

**C. Condition-Invariance Methods:** CNN-based methods for condition-invariant VPR leverage unsupervised learning, context-aware adaptation, patch-level aggregation, and descriptor mixing to enhance robustness across environmental variations. Recent advances further incorporate viewpoint-invariant modeling, large-scale supervised training, and domain-aligned feature aggregation, collectively improving generalization under diverse real-world conditions. The detailed description is shown in Table 3.

The methods summarized in Table 3 represent a range of condition-invariant strategies within VPR, each addressing environmental variability through distinct architectural or learning mechanisms. Techniques such as Appearance-based VPR and CRN explicitly target robustness by leveraging unsupervised feature learning and context-aware reweighting, respectively, enabling adaptability to changes in illumination and appearance. Patch-NetVLAD [57] and MixVPR [58] enhance local and global feature robustness via patch-level aggregation and multi-scale feature mixing. EigenPlace [59] and CoSTPlace [60] improve invariance to viewpoint and urban complexity through viewpoint-invariant descriptors and large-scale classification training. SALAD [19] introduces opti-mal transport for feature aggregation, promoting domain generalization across diverse conditions.

Table 3: CNN-based Condition-Invariant VPR Methods

| Method | Key Features | Condition-Invariance Justification |
|---|---|---|
| Appearance-based VPR [47] | Unsupervised features for condition-robust recognition | Designed specifically for condition robustness |
| CRN [50] | Context-aware local feature reweighting | Adapts features based on environmental context |
| Patch-NetVLAD [57] | Patch-level aggregation from NetVLAD residuals | Enhances robustness to local appearance changes |
| MixVPR [58] | Feature mixing for compact global descriptors | Combines feature representations to handle varied conditions |
| EigenPlaces [59] | Viewpoint-invariant global descriptor learning | Explicitly targets viewpoint variation |
| CosPlace [60] | Classification-based training with city-scale dataset | Learns robust features across diverse urban conditions |

**Remark 2:** *CNN-based methods for VPR are highly effective because they automatically learn discriminative hierarchical features directly from image data, providing robustness to variations in viewpoint, lighting, and environmental conditions. Methods such as NetVLAD, GeM, and MixVPR excel at aggregating global features for accurate place recognition, particularly in large-scale and complex environments. However, they are computationally intensive, prone to overfitting with limited data, and may struggle with capturing fine-grained local features. Techniques like SPE-VLAD and Patch-NetVLAD enhance performance by incorporating multi-scale and local feature aggregation. Despite these advancements, CNN-based methods still depend heavily on large labeled datasets and may underperform in data-scarce situations unless augmented with self-supervised or unsupervised learning strategies.*

*3.1.4. Transformer-Based VPR Methods.*

The Transformer is a deep learning architecture based on the self-attention mechanism, which initially achieved remarkable success in natural language processing (NLP) tasks. Compared to traditional convolutional neural networks (CNNs), the greatest advantage of the Transformer is its ability to model global context, meaning it can effectively capture long-range dependencies in input data. This makes the Transformer excel in tasks involving global features of images, especially in visual place recognition tasks under complex scenes. The global/local feature-based VPR method can be described as shown in Figure 10, and the detail category is shown in Table 4.

**A. Global Descriptor Methods:** Global descriptor methods focus on generating a single, fixed-size feature vector that represents the entire image.

TransVPR [62] represents a significant advancement in VPR by effectively combining the global contextual understanding of Vision Transformers with the precision of key-patch descriptors, offering a robust and efficient solution for real-world localization challenges. DINOv2 [63] employs a self-distillation ap-proach, where a student model learns to predict the output of an exponentially moving average of its previous states, serving as the teacher, which enables the model to learn robust representations without labeled data. TransVLAD [65] employs a sparse transformer to encode global dependencies and compute attention-based feature maps to effectively reduce visual ambiguities that occur in large-scale geo-localization problems, enhancing the model's ability to handle diverse visual cues. BoQ [67] leverages a set of learnable global queries to probe local features via cross-attention, ensuring consistent information aggregation across varying environmental conditions and viewpoints. SelaVPR [69] employs lightweight adapters to adapt pre-trained models without modifying their core parameters. This approach facilitates the extraction of both global and local features, focusing on salient landmarks for accurate place discrimination. ProGEO [70] leverages the multi-modal capabilities of CLIP to create a set of learnable text prompts for each geographic image feature. These prompts form vague descriptions that assist in aligning visual features with semantic information, enhancing the model's understanding of the image context. EffoVPR [72] utilizes features extracted from the self-attention layers of DINOv2 as a powerful re-ranking mechanism, which allows for effective zero-shot retrieval, outperforming previous methods that relied solely on global features. Pair-VPR [20] represents a significant advancement in VPR by effectively integrating pre-training with masked image modeling and contrastive learning, offering a robust and efficient solution for real-world localization tasks. SelaVPR++ [74] employs lightweight multi-scale convolution (MultiConv) adapters to refine intermediate features from a frozen foundation model. This approach avoids backpropagating gradients through the backbone during training, significantly reducing computational overhead. SALAD [19] introduces a novel approach to VPR by reformulating the feature aggregation process as an optimal transport problem. This method aims to enhance the quality of global image descriptors by effectively assigning local features to clusters and discarding non-informative features. AnyLoc [75] leverages general-purpose feature representations from off-the-shelf self-supervised models, such as DINOv2, without any VPR-specific training. This method combines these features with unsupervised feature aggregation techniques, like VLAD and GeM pooling, to create robust global descriptors.

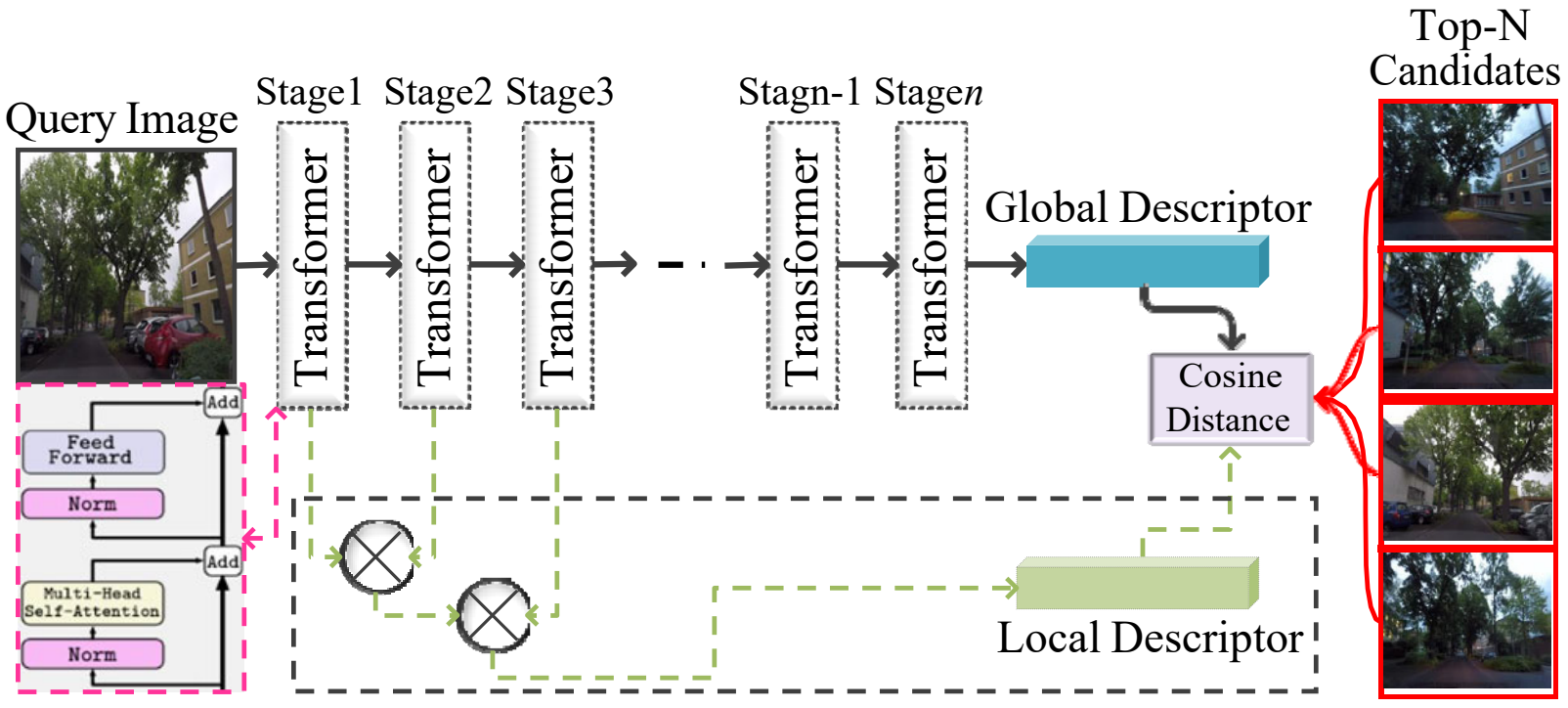

Figure 10: Transformer-based VPR paradigm. The figure describes two different feature representation methods: the method based on local features and the method based on global features.

Table 4: Transformer-based VPR Methods.

| Method | Published In | Key Features | Type |
|---|---|---|---|
| TransVPR [62] | CVPR 2022 | ViT with multi-level attention for global/local fusion | Global-based |
| DINOv2 [63] | ArXiv 2023 | Self-distillation with ViT backbones | Global-based |
| Superglue [64] | CVPR 2020 | GNN for local feature matching | Local-based |
| TransVLAD [65] | WACV 2023 | Transformer-enhanced VLAD aggregation | Global-based |
| DHE [66] | AAAI 2024 | Transformer for direct homography estimation | Local-based |
| BoQ [67] | CVPR 2024 | Bag of learnable queries with attention | Global-based |
| CricaVPR [68] | CVPR 2024 | Cross-image attention-based representation learning | Local-based |
| SelaVPR [69] | ArXiv 2024 | Adapter-based transformer fine-tuning | Global-based |
| ProGEO [70] | ICANN 2024 | Prompt generation using CLIP features | Global-based |
| SegVLAD [71] | ECCV 2024 | Segment-level retrieval using VLAD | Local-based |
| EffoVPR [72] | ArXiv 2024 | DINOv2 self-attention re-ranking | Global-based |
| Pair-VPR [20] | RAL 2025 | Pretraining + pair classification via ViT | Global-based |
| PRGS [73] | PR 2025 | Patch-region graph search for reranking | Local-based |
| SelaVPR++ [74] | ArXiv 2025 | MultiConv adapters for efficient foundation model adaptation | Global-based |
| SALAD [19] | CVPR 2024 | Optimal transport for feature aggregation | Global-based |
| AnyLoc [75] | RAL 2023 | Unsupervised feature use from general models | Global-based |

**B. Local Descriptor Methods:** Local descriptor methods in VPR focus on extracting and matching local features from images to identify places. These methods often emphasize robustness to local environmental variations, such as changes in viewpoint, lighting, and occlusion.

Superglue [64] revolutionizes the process of local feature matching by employing a graph neural network (GNN) to jointly establish correspondences and reject non-matching points, addressing challenges such as occlusion, viewpoint changes, and illumination variations. DHE [66] presents a transformer-based DHE network that takes dense feature maps extracted by a backbone network as input, which directly fits homography for fast and learnable geometric verification, eliminating the need for traditional RANSAC-based methods. CricaVPR [68] utilizes a self-attention mechanism to model the relationships between multiple images within a batch, which allows the model to capture variations in viewpoint and illumination, leading to more robust feature representations. SegVLAD [71] proposes encoding and retrieving image segments—distinct, meaningful parts of an image—rather than whole images. Using open-set image segmentation, SegVLAD decomposes images into 'things' and 'stuff', creating a representation called SuperSegment. PRGS [73] constructs a graph representation of image patches and their spatial relationships, facilitating the identification of semantically consistent regions across different views, which enables the model to focus on region-level correspondences, improving the accuracy of place recognition under varying conditions.

**C. Reranking Methods:** Reranking methods in Transformer-based VPR play a crucial role in refining the initial retrieval results by focusing on enhancing the relationships between query and reference images, improving the final accuracy of place recognition systems. These methods, which use attention mechanisms or cross-image feature relationships, are especially valuable in complex and large-scale datasets where initial retrieval might not always be optimal.

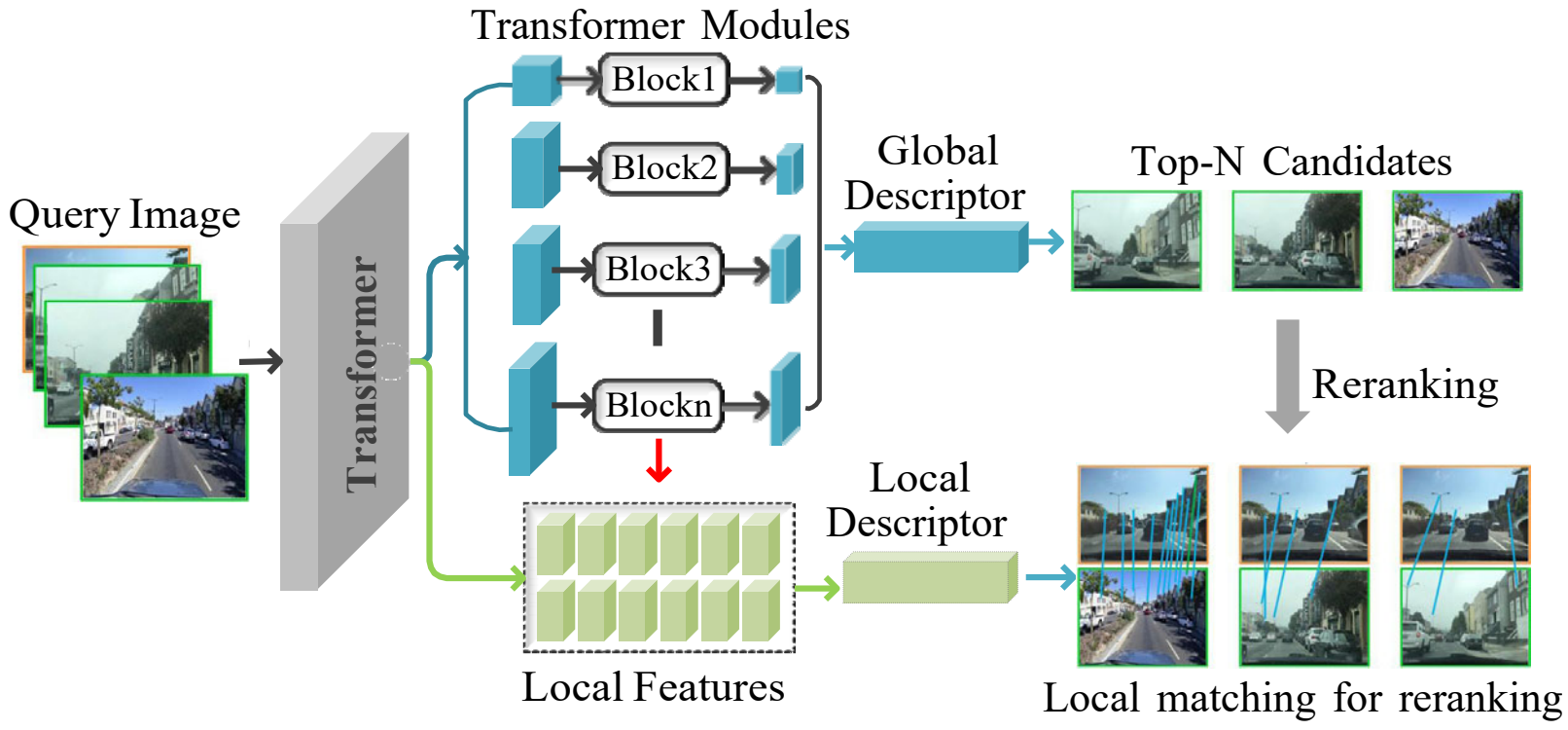

Figure 11: Reranking-based VPR paradigm. Reranking-based place recognition enhances the accuracy of place recognition systems by integrating an initial retrieval phase (global retrieval) with a subsequent reranking step (local retrieval).

Reranking-based VPR paradigm can be described as shown in Figure 11.

Reranking methods also can be classified as: 1) Attention-Based Reranking: Several methods like TransVPR, EffoVPR, R2Former, and BoQ focus on using the attention mechanism to refine the ranking by adjusting the importance of various image features, improving accuracy in diverse environments. 2) Cross-Attention and Feature Aggregation: BoQ uses cross-attention to probe local features with global queries, while CricaVPR models relationships between images to rerank based on contextual relevance. 3) Feature Matching Refinement: SelaVPR and ProGEO improve initial matches by leveraging learned feature relations (either local or semantic) to fine-tune the retrieval ranking. The detail category is shown in Table 5.

Table 5: Reranking Categories of Transformer-based VPR Methods.

| Method | Reranking Category | Key Mechanism Description |
|---|---|---|
| TransVPR [62] | Attention-Based Reranking | Multi-level attention mechanism |
| EifoVPR [72] | Attention-Based Reranking | DINOv2 self-attention reranking |
| BoQ [67] | Attention + Cross-Attention | Learnable queries with cross-attention |
| CricaVPR [68] | Cross-Attention and Feature Aggregation | Modeling contextual relationships between images |
| SelaVPR [69] | Feature Matching Refinement | Transformer fine-tuning for match refinement |
| ProGEO [70] | Feature Matching Refinement | Prompt generation with semantic match optimization |
| R2Former [76] | Attention-Based Reranking | jointly models retrieval and reranking within a single framework |

**Remark 3:** *Transformer-based methods in Visual Place Recognition (VPR) utilize Vision Transformers (ViTs) and attention mechanisms to capture global context and local features, offering advantages in handling complex, large-scale localization tasks. Methods like TransVPR, DINOv2, and BoQ excel at learning long-range dependencies and improving robustness to environmental variations. However, they are computationally intensive and require large, well-annotated datasets for optimal performance. Despite these challenges, transformer-based approaches have demonstrated superior accuracy and scalability, particularly in diverse and dynamic environments, making them increasingly popular for advanced VPR tasks.*

*3.1.5. Other Categories*

Some methods do not have clear global features or local feature extraction, some may be based on probabilistic or mixed methods, which are shown in Table 6. we classify these methods into one category. For example, Methods such as HMM, Hiervpr, and MRS-VPR do not use deep learning frameworks. Usually, they are based on probabilistic models, image processing methods, or optimization algorithms for matching and recognition. These methods, such as NetVLAD, DINO, TransVPR, etc., rely on deep neural networks and use architectures such as CNN or Transformer to learn features and perform visual place recognition.

HMM [77] integrates Hidden Markov Models (HMMs) with visual sequence

Table 6: Other categories

| Method | Published In | Key Features | Type |
|---|---|---|---|
| HMM [77] | IROS 2014 | Hidden Markov Models with visual sequence matching for dynamic environments. | Traditional Method |
| Hiervpr [78] | ICRA 2015 | Hierarchical matching framework integrating environment-specific utilities. | Traditional Method |
| MRS-VPR [42] | ICCV 2019 | Multi-resolution approach for sequence-based place recognition. | Traditional Method |
| AP-GEM [79] | ICCV 2019 | Optimizes mean Average Precision (mAP) for training image retrieval systems. | CNN-based |
| SARE [80] | ICCV 2019 | Loss function capturing intra-place and inter-place relationships. | CNN-based |
| DINO [81] | ICCV 2021 | Self-distillation without labels, matching outputs of student and teacher networks. | Transformer-based |
| HAF [82] | ICASSP 2021 | CNN for hierarchical feature extraction, capturing both fine-grained and global context. | CNN-based |
| GCL [83] | ArXiv 2021 | Generalized contrastive loss for Siamese networks, improving VPR performance. | CNN-based |

matching to improve place recognition, especially in dynamic or partially observable environments. Hiervpr [78] represents a significant advancement in VPR by integrating environment and place-specific utilities into a hierarchical matching framework, enhancing both accuracy and efficiency in diverse and challenging environments. MRS-VPR [42] introduces a multi-resolution approach to sequence-based place recognition. Instead of matching image sequences at a single resolution, MRS-VPR processes images at multiple resolutions (scales) to improve matching performance. AP-GEM [79] introduces a novel approach to training image retrieval systems by directly optimizing the mean Average Precision (mAP) metric through a differentiable, listwise loss function. SARE [80] presents a significant advancement in the field of image-based localization by introducing a loss function that effectively captures the nuances of intra-place and inter-place relationships, leading to improved performance in large-scale scenarios. DINO [81] operates as a form of self-distillation without labels, where a student network learns to match the output of a teacher network through a cross-entropy loss. The teacher's parameters are updated using an exponential moving average of the student's parameters, and techniques like centering and sharpening are applied to prevent collapse. HAF [82] utilizes CNN to extract hierarchical feature maps at multiple levels, capturing both fine-grained details and global context. GCL [83] introduces a novel approach to training Siamese CNNs for VPR, which addresses the limitations of traditional binary contrastive loss functions by proposing a Generalized Contrastive Loss GCL that incorporates continuous similarity measures between image pairs.

**Remark 4:** *Methods outside the CNN-based and Transformer-based categories in VPR typically rely on traditional machine learning techniques, probabilistic models, or hybrid frameworks. Approaches such as HMM, Hiervpr, and MRS-VPR utilize sequence-based, hierarchical, or multi-resolution strategies to enhance place recognition. While effective in dynamic or scale-varying environments, these methods often require manual feature extraction and are computationally intensive. Techniques like AP-GEM and SARE optimize image retrieval and geo-localization performance through advanced loss functions but may lack the flexibility and scalability of deep learning approaches. While non-deep learning methods offer valuable solutions in specific contexts, they are* *limited by their reliance on traditional methods and do not fully leverage the end-to-end learning capabilities and scalability of CNN-based or Transformer-based methods.*

### *3.2. Lidar Place Recognition (LPR)*

LPR enables autonomous vehicles to localize by matching current scans with prior LiDAR data, solving loop closure and global localization. Its robustness to illumination and appearance changes makes it essential for long-term, large-scale navigation. This process addresses two key problems: loop closure detection—answering "Where have I ever been?"—to ensure map consistency, and global localization—answering "Where am I?". The detailed process is illustrated in Figure 12. As shown in Figure 13, the evolution of LPR methods is demonstrated.

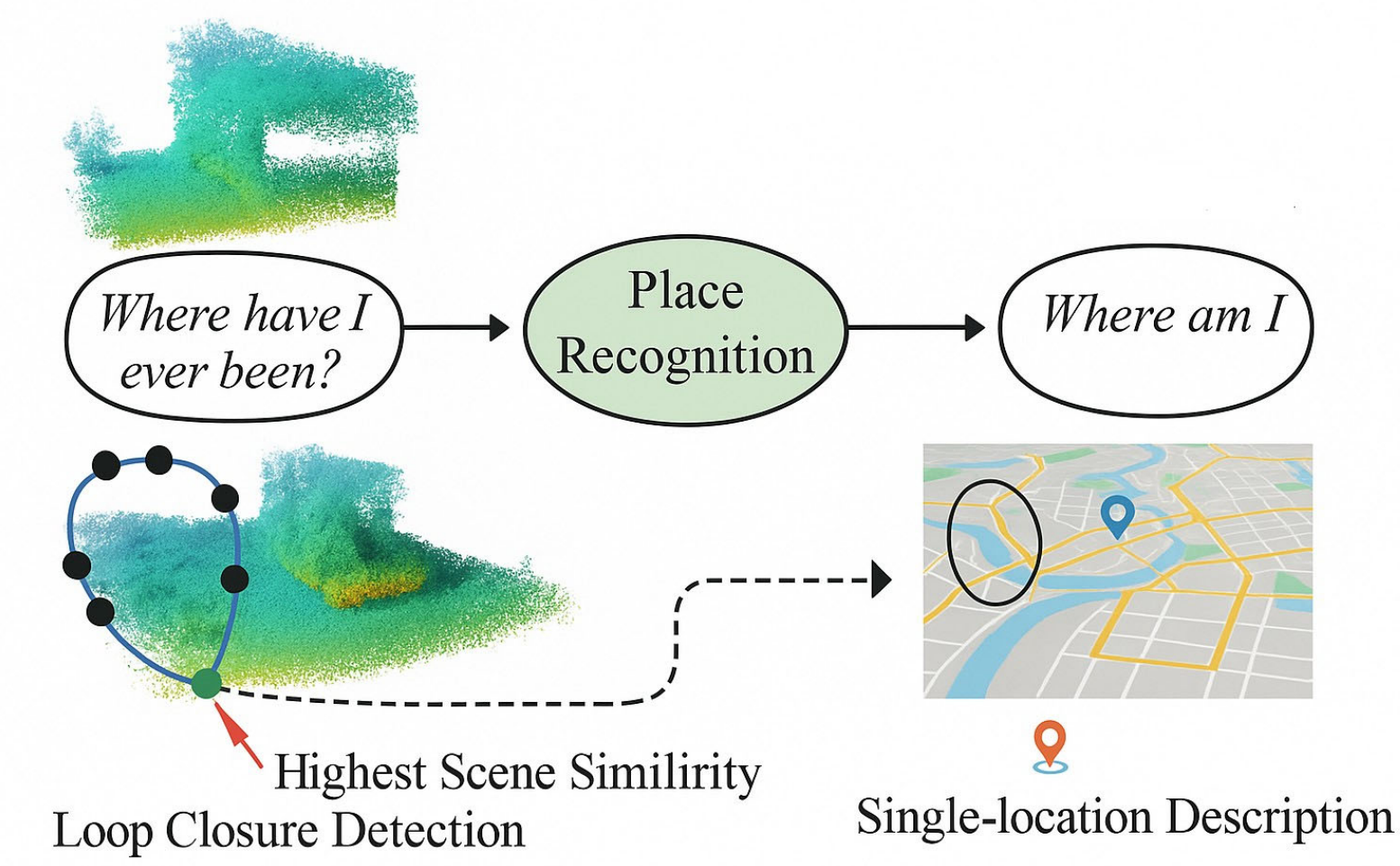

Figure 12: PR addresses two key issues: On the left, blue lines represent vehicle trajectories, while solid circles indicate scans collected by sensors over time. On the right, the highlighted area illustrates the global distribution of vehicles, offering only a single location description on the map.

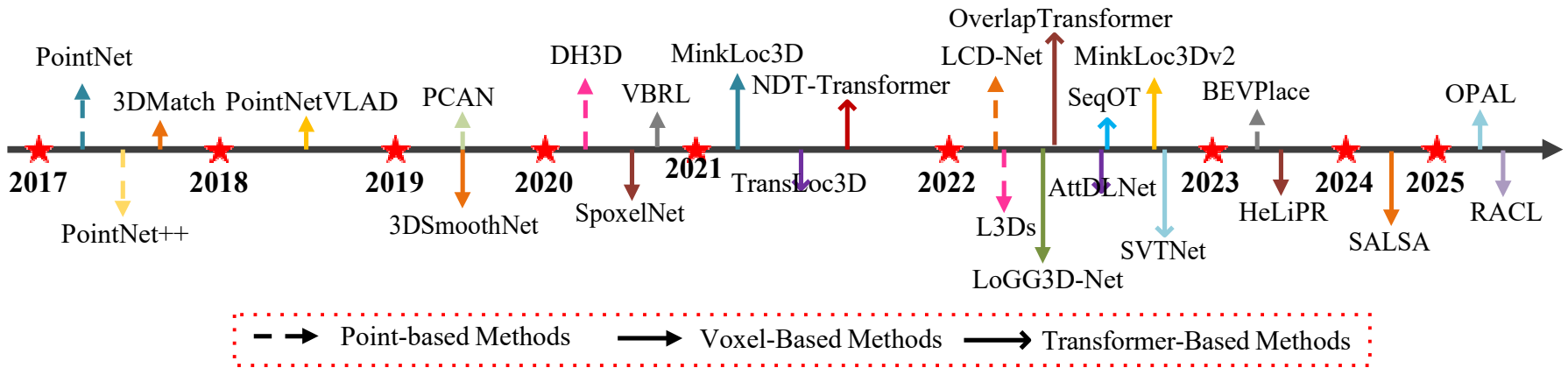

Figure 13: The different mainstream methods in the field of LPR in the recent decade. According to the point encoding method, LPR methods are subdivided into point-based methods, voxel-based methods, and Transformer-based methods.

### *3.2.1. Point-Based Methods*

Point-Based VPR Methods extract features directly from raw 3D point clouds, preserving fine-grained geometry without voxelization. Starting with PointNet and PointNet++, these methods evolved through attention mechanisms and global aggregation (e.g., NetVLAD) to enhance place discriminability. They offer high accuracy in complex environments, but face challenges in scalability and efficiency on dense data. The point-based LPR paradigm is shown in Figure 14. The summary of point-based LPR methods is shown in Table 7.

PointNet [84] introduces a groundbreaking architecture based on multi-layer perceptrons (MLPs) that directly processes unordered 3D point sets, effectively capturing global geometric structures while ensuring permutation invariance. PointNet++ [85] enhances local geometric awareness through hierarchical feature aggregation, allowing the model to learn both global and fine-grained spatial patterns. PointNetVLAD [86] integrates PointNet with the NetVLAD aggregation module to produce compact global descriptors for large-scale place recognition from raw point clouds. PCAN [87] incorporates attention mechanisms into point cloud processing, selectively highlighting informative regions to enhance discriminative capabilities in challenging scenarios. DH3D [88] unifies local feature detection and description through point convolution within a single-pass framework, facilitating efficient end-to-end six degrees of freedom (6DoF) place recognition. LCD-Net [89] utilizes a point-voxel hybrid representation to capture both fine local details and robust global context, thereby improving performance in loop closure detection. L3Ds [90] combines the structural encoding of PointNet with Transformer attention modules, providing robust place recognition in cluttered and dynamic environments. SOLiD [91] is a compact point-based place recognition framework designed to handle field-of-view (FoV) limitations. It leverages a spatial reorganization scheme and encodes height-directional attention to improve recognition under constrained perspectives, such as narrow LiDAR views.

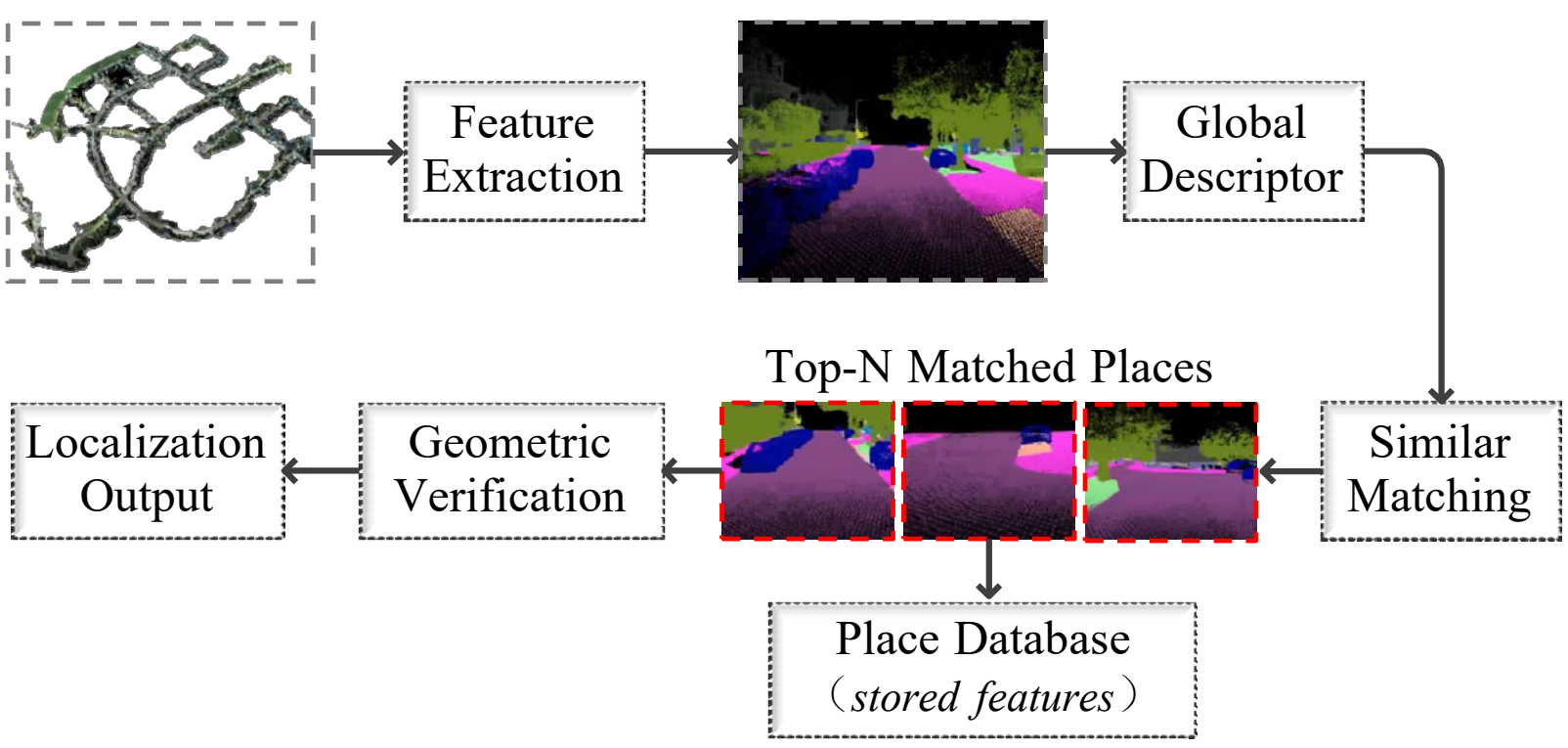

Figure 14: Point-based LPR uses point cloud features to generate global descriptors, match candidate places from a database, and verify geometry for accurate localization in 3D environments.

**Remark 5:** *Point-based VPR methods, such as PointNetVLAD and PCAN, preserve fine-grained geometry from raw point clouds and offer strong discriminative power through global aggregation and attention mechanisms. While they achieve high accuracy in complex scenes, these methods face scalability issues with dense data and reduced robustness under occlusion or viewpoint changes. Hybrid models like L3Ds alleviate some limitations, but efficiency and adaptability remain challenges in real-world applications.*

#### *3.2.2. Voxel-Based Methods*

Voxel-based LPR methods operate by discretizing raw 3D point clouds into structured voxel grids, enabling the use of convolutional architectures for efficient feature extraction. This transformation converts irregular, unordered point data into regular 3D tensors, where each voxel encodes spatial statistics such as occupancy, density, height, or intensity. The voxel-based LPR paradigm is shown in Figure 15. The summary of voxel-based LPR methods is shown in Table 8.

Table 7: Summary of Point-Based VPR Methods.

| Method | Published In | Key Features | Type |
|---|---|---|---|
| PointNet [84] | CVPR 2017 | End-to-end cloud processing | MLP |
| PointNet++ [85] | NeurIPS 2017 | Local structure+hierarchical structure | MLP |
| PointNetVLAD [86] | CVPR 2018 | PointNet+NetVLAD | MLP+Aggregation Layer |
| PCAN [87] | CVPR 2019 | The attention mechanism enhances local features | MLP + Attention |
| DH3D [88] | ECCV 2020 | Multi-scale point convolution, FlexConv | PointConv |
| LCD-Net [89] | TRO 2022 | Point-voxel mixed structure, aggregating global features | PV-RCNN |
| L3Ds [90] | RAL 2022 | PointNet + Transformer | Hybrid |
| SOLiD [91] | RAL 2024 | Spatial reorganization, Height-directional encoding | Point-based |

3DMatch [92] introduces a voxel-based local descriptor by applying 3D

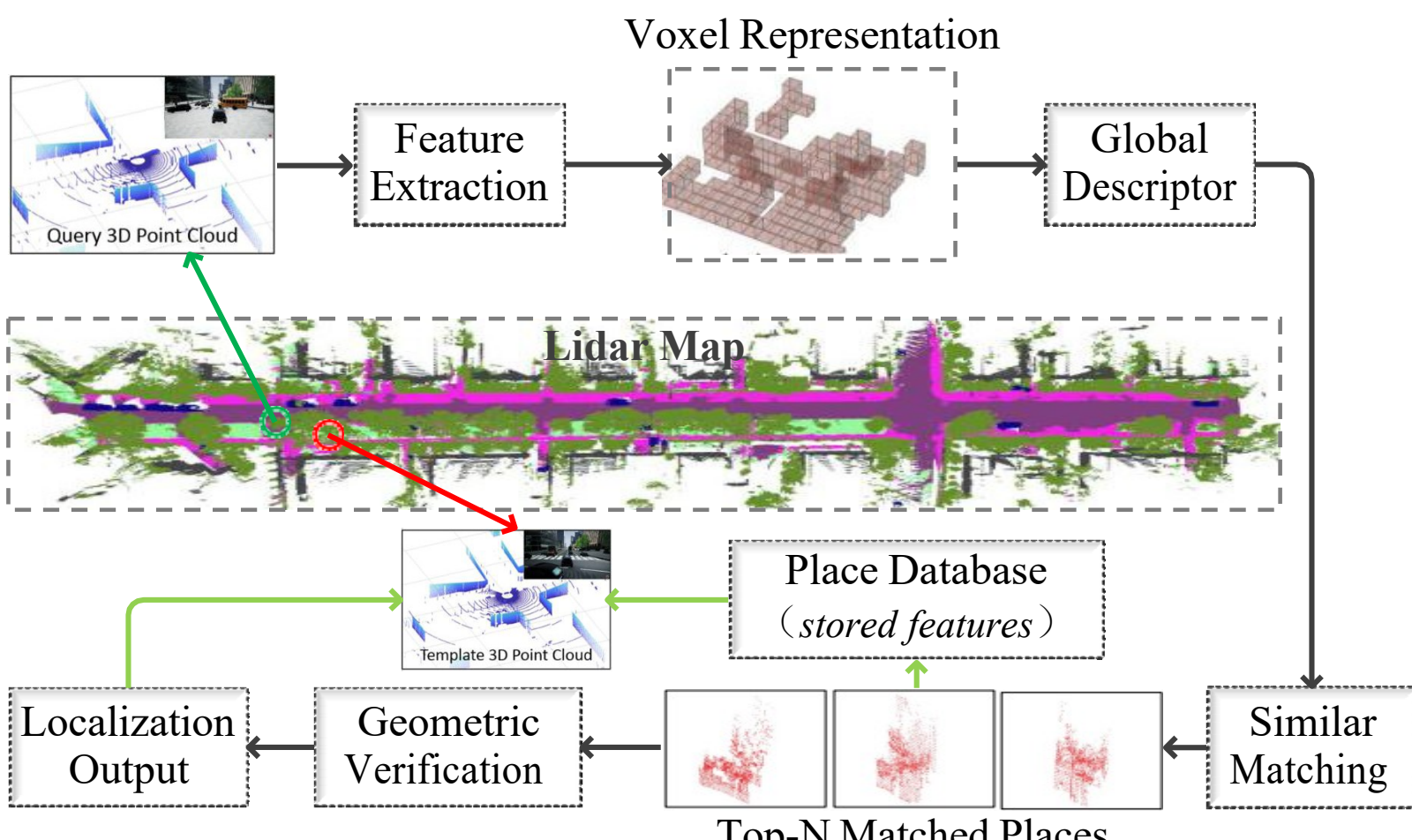

Figure 15: Voxel-based LPR converts raw point clouds into voxel grids, extracts structured geometric features, generates a global descriptor, matches it with a stored place database, and verifies the match geometrically to achieve robust 3D localization.

CNNs to Truncated Signed Distance Function (TSDF) voxel grids. This method captures local geometric patterns, facilitating robust matching under varying viewpoints and occlusions, and serves as a foundational technique in volumetric place recognition. 3DSmoothNet [93] enhances 3D voxel descriptors by learning smoothed density representations, which improve rotation invariance and generalization. It focuses on extracting repeatable local features using fixed voxel neighborhoods, thereby enabling reliable correspondences in cluttered scenes. SpoxelNet [94] employs a spherical coordinate system for voxelization and introduces multi-scale stitching to encode geometric structures. This approach allows for robust descriptor learning under occlusion and varying perspectives, while efficiently representing global features using dense neural networks. VBRL [95] integrates sparse regularization with multimodal feature fusion, aiming to enhance long-term place recognition performance. It introduces structured sparsity constraints to jointly optimize voxel importance and modality-specific features. MinkLoc3D [96] leverages sparse CNNs combined with GeM pooling to efficiently generate global descriptors from voxelized LiDAR data. Its architecture enables high-speed inference while maintaining robust discriminative capacity across diverse scenes. MinkLoc3Dv2 [97] improves upon its predecessor by refining the loss function and applying advanced descriptor supervision strategies, which enhance retrieval accuracy and training stability. It further strengthens descriptor distinctiveness for large-scale place recognition. LoGG3D-Net [98] introduces a point-voxel hybrid convolution framework with local geometric consistency constraints. This framework promotes spatial alignment between frames and improves robustness to temporal drift and dynamic object interference, offering strong performance in real-world scenarios. BEVPlace [99] projects LiDAR scans into bird's eye view (BEV) representations, capturing spatial context in a 2D grid format. By applying CNNs to these structured views, BEVPlace achieves viewpoint-invariant place recognition, suitable for long-term localization tasks with large environmental variations. The HeLiPR dataset [100] introduces benchmark methods operating across heterogeneous LiDAR sensors (e.g., VLP-16, Livox Avia). The baseline models adopt voxel grid representations to evaluate robustness to varying LiDAR characteristics and cross-device generalization, a critical but underexplored challenge in place recognition.

Table 8: Summary of Voxel-Based VPR Methods.

| Method | Published In | Key Features | Type |
|---|---|---|---|
| 3DMatch [92] | CVPR 2017 | TSDF Voxel coding+CNN | Voxel CNN |
| 3DSmoothNet [93] | CVPR 2019 | Smooth density voxel representation | Voxel CNN |
| SpoxelNet [94] | IROS 2020 | Spherical coordinate voxel coding, Multi-scale stitching | DNN |
| VBRL [95] | IROS 2020 | Sparse regularization+Multimodal fusion | Hybrid |
| MinkLoc3D [96] | WACV 2021 | Sparse voxel convolution+GeM Pooling | Sparse CNN |
| MinkLoc3Dv2 [97] | ICPR 2022 | Improve the loss function and enhance the discrimination ability | Sparse CNN |
| LoGG3D-Net [98] | ICRA 2022 | Point voxel convolution, Local consistency optimization | Sparse CNN |
| BEVPlace [99] | ICCV 2023 | BEV representation, Viewpoint invariance | Voxel-based |
| HeLiPR [100] | IJRR 2024 | Cross-device voxel grid evaluation, Heterogeneous LiDAR setup | Voxel-based |

**Remark 6:** *Voxel-based LPR methods convert raw point clouds into structured voxel grids, facilitating efficient feature extraction through convolutional architectures. Foundational approaches, such as 3DMatch and 3DSmoothNet, concentrate on local geometric patterns and rotation-invariant descriptors. In contrast, SpoxelNet and VBRL improve robustness by employing spherical voxelization and multimodal fusion techniques. Recent models, including MinkLoc3D and LoGG3D-Net, utilize sparse or hybrid voxel frameworks combined with pooling and consistency constraints, achieving high efficiency and accuracy in large-scale applications. Despite their advantages, voxel-based methods encounter challenges, including discretization loss and memory overhead in high-resolution environments.*

#### *3.2.3. Transformer-Based Methods*

Transformer-based PR methods have recently advanced LiDAR place recognition by effectively capturing both spatial and temporal dependencies in point clouds. The Transformer-based LPR paradigm is shown in Figure 16. The summary of Transformer-based LPR methods is shown in Table 9.

TransLoc3D [101] introduces adaptive receptive fields that dynamically adjust feature extraction based on scene complexity, thereby enhancing the robustness of global descriptors. NDT-Transformer [102] utilizes the Normal Distribution Transform within a Transformer encoder to model local spatial distributions, which improves localization in GPS-denied or repetitive environ-ments. HiTPR [103] employs a hierarchical architecture with dual-level attention to jointly learn short-range geometric and long-range contextual features, facilitating generalization across various viewpoints. OverlapTransformer [104] addresses viewpoint sensitivity by incorporating a yaw-invariant Transformer and dense attention mechanisms for overlap prediction, excelling in reverse-loop scenarios. AttDLNet [105] utilizes a lightweight attention framework to encode global spatial context, providing efficiency that is suitable for real-time applications. SeqOT [106] integrates spatial-temporal cues from sequential range images to achieve viewpoint-invariant recognition in dynamic scenes. Finally, SVT-Net [107] combines sparse voxel representations with lightweight Transformer modules, striking a balance between recognition accuracy and computational efficiency, making it ideal for large-scale deployment. SALSA [108] proposes a radial attention mechanism within a Sphereformer backbone to learn global representations of sparse 3D point clouds. The architecture combines local radial-window self-attention and feed-forward mixing, offering a scalable and interpretable framework for global localization in large-scale environments. OPAL [109] is a multimodal Transformer-based method that fuses LiDAR scans with topological data from OpenStreetMap (OSM). It introduces visibility mask-ing and adaptive radial fusion layers to integrate geometric and semantic cues, addressing cross-modality matching in urban-scale environments. RACL [110] targets lifelong place recognition by incorporating a ranking-aware continual learning framework. A Transformer-based backbone is enhanced with memory replay and distributional alignment techniques to preserve discriminative descriptors across incremental training episodes.

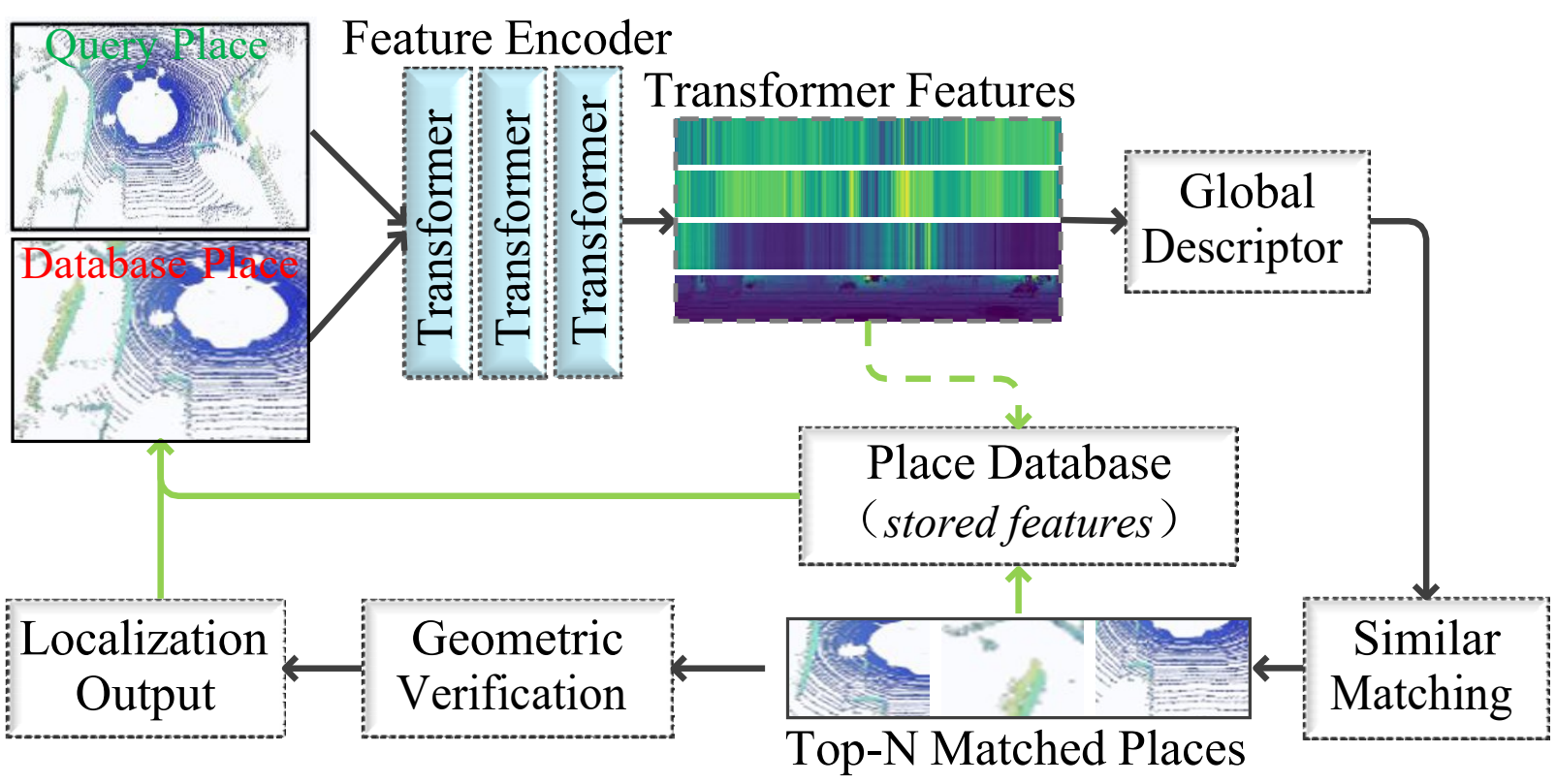

Figure 16: Transformer-based LPR method uses Transformer networks to model long-range dependencies and spatial relationships within LiDAR point clouds, enabling discriminative global descriptors for accurate place recognition.

**Remark 7:** *Transformer-based LPR methods, such as TransLoc3D, HiTPR, and NDT-Transformer, effectively capture spatial and temporal dependencies in LiDAR data, enhancing robustness to viewpoint changes, structural variation, and GPS-denied conditions. Lightweight models like AttDLNet and SVT-Net offer improved efficiency for real-time use. While these methods excel in global context modeling and generalization, they remain limited by high data and computation demands, posing challenges for deployment in sparse or dynamic environments.*

Table 9: Summary of Transformer-Based VPR Methods.

| Mehtod | Published In | Key Features | Type |
|---|---|---|---|
| TransLoc3D [101] | ArXiv 2021 | Multi-scale Transformer+Attention aggregation | Transformer |
| NDT-Transformer [102] | IROS 2021 | NDT structure+Transformer encoder | Transformer |
| HiTPR [103] | ICRA 2022 | Multi-layer Transformer extracts local and global contexts | Transformer |
| OverlapTransformer [104] | RAL 2022 | Dense registration, adapting to reverse viewing angles and occlusions | Transformer |
| AttDLNet [105] | Iberian Robotics conference 2022 | Attention network models scene relationships | Transformer |
| SeqOT [106] | TIE 2022 | Multi-scale Transformer processes LiDAR sequences | Transformer |
| SVTNet [107] | AAAI 2022 | Sparse voxel Transformer, lightweight structure | Transformer |
| SALSA [108] | RAL 2024 | Radial attention with Sphereformer backbone, Sparse point encoding | Transformer |
| OPAL [109] | ArXiv 2025 | Multimodal fusion, Adaptive radial fusion and Visibility masking | Transformer |
| RACL [109] | ArXiv 2025 | Continual learning, Ranking preservation, Memory-based training | Transformer |

### *3.3. Cross-Modal Place Recognition (CMPR)*

CMPR aims to identify whether two observations acquired from different modalities (e.g., text-Lidar, text-image) correspond to the same physical location. Unlike unimodal place recognition, which relies on intra-modal feature similarities, CMPR must overcome domain gaps such as appearance-geometry discrepancies, modality-specific noise, and inconsistent viewpoints. As shown in Fig. 17, the evolution of CMPR methods is demonstrated. The CMPR paradigm (including text to image, text to Lidar) is illustrated in Figure 18.

The summary of Transformer-based LPR methods is shown in Table 10.

**A. Text-Lidar LPR Methods:** Text-Lidar methods refer to a class of

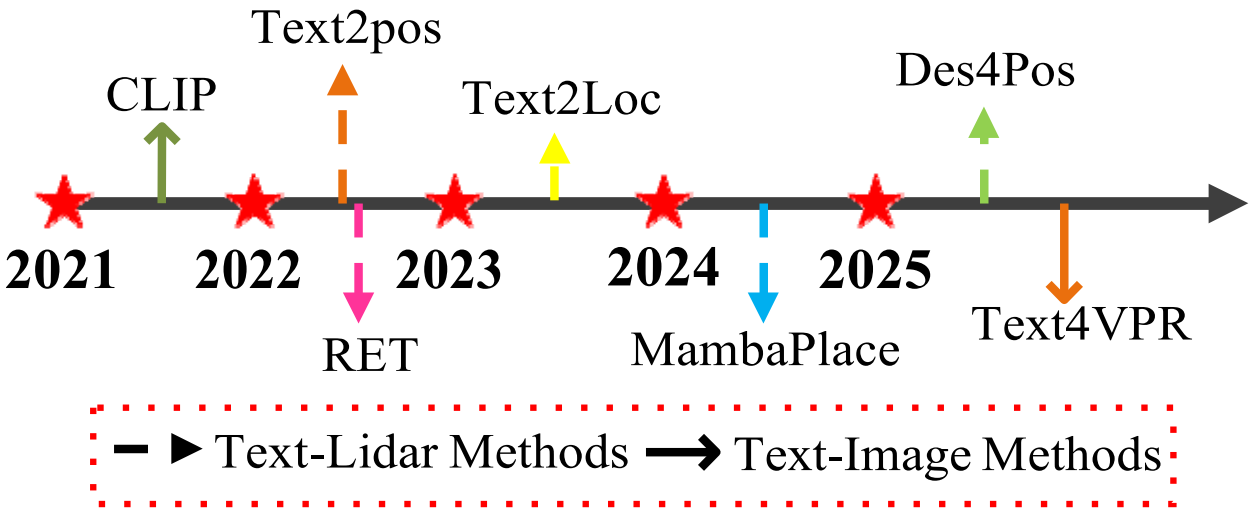

Figure 17: The mainstream methods in the field of CMPR in recent years. According to the fused modalities, CMPR methods are subdivided into text-Lidar methods, text-image methods.

cross-modal approaches that aim to bridge natural language and Lidar-based 3D representations. These methods enable machines to interpret, retrieve, or localize 3D spatial environments based on textual descriptions, and have growing applications in robotics, autonomous driving, and embodied AI.

Table 10: Summary of CMPR Methods.

| Method | Published In | Key Features | Type |
|---|---|---|---|
| CLIP [111] | ICML 2021 | Contrastive vision-language pretraining on large-scale image-text pairs | Text-Image, Pretraining |
| RET [23] | AAAI 2023 | Relation-enhanced Transformer for explicit text-to-point relationships | Text-Lidar, Transformer-based |
| Text2Pos [22] | CVPR 2022 | Two-stage localization: coarse retrieval+6-DoF pose regression | Text-Lidar, Regression-based |
| Text2Loc [21] | CVPR 2024 | Hierarchical Transformer with frozen T5 and contrastive fine-tuning | Text-Lidar, Transformer-based |
| MambaPlace [112] | ArXiv 2024 | Mamba-based SSM + Attention, coarse-to-fine place recognition framework | Text-Lidar, SSM + Attention |
| Des4Pos [113] | ArXiv 2025 | Bi-LSTM with multi-scale attention for sparse LiDAR scenes | Text-Lidar, RNN-based |
| Text4VPR [114] | CVPR 2025 | Sinkhorn alignment + multi-view cross-attention for image-text matching | Text-Image, Registration-based |

RET [23] proposes a Relation-Enhanced Transformer to explicitly model semantic relationships between language queries and candidate objects in 3D point clouds. This method effectively captures fine-grained correspondences and improves grounding accuracy by integrating linguistic context with geometric structure, making it well-suited for referential localization tasks. Text2Pos [22] is the first approach to directly regress a 6-DoF camera pose from natural language descriptions within 3D point clouds. The method employs a two-stage architecture—coarse submap retrieval followed by fine-grained pose regression—thereby bypassing traditional retrieval pipelines and enabling precise cross-modal local-ization through textual input alone. Text2Loc [21] introduces a hierarchical Transformer-based model that encodes textual and 3D data without relying on explicit instance grounding. It utilizes a frozen T5 language encoder and contrastive learning for coarse localization, followed by fine-level refinement via attention. The model demonstrates improved robustness to ambiguous or imprecise descriptions compared to prior methods. MambaPlace [112] leverages a selective state space model (Mamba) combined with attention-based modules in a coarse-to-fine dual-branch framework. It enhances cross-modal representation learning by capturing long-range dependencies and structured scene semantics, achieving state-of-the-art text-driven 3D place recognition performance. Des4Pos [113] targets the autonomous driving setting and proposes a language-conditioned place recognition framework based on bidirectional LSTM encoders and multi-scale attention for LiDAR point clouds. It effectively captures both local geometric structures and global semantic cues from textual queries, demonstrating high accuracy in sparse 3D environments.

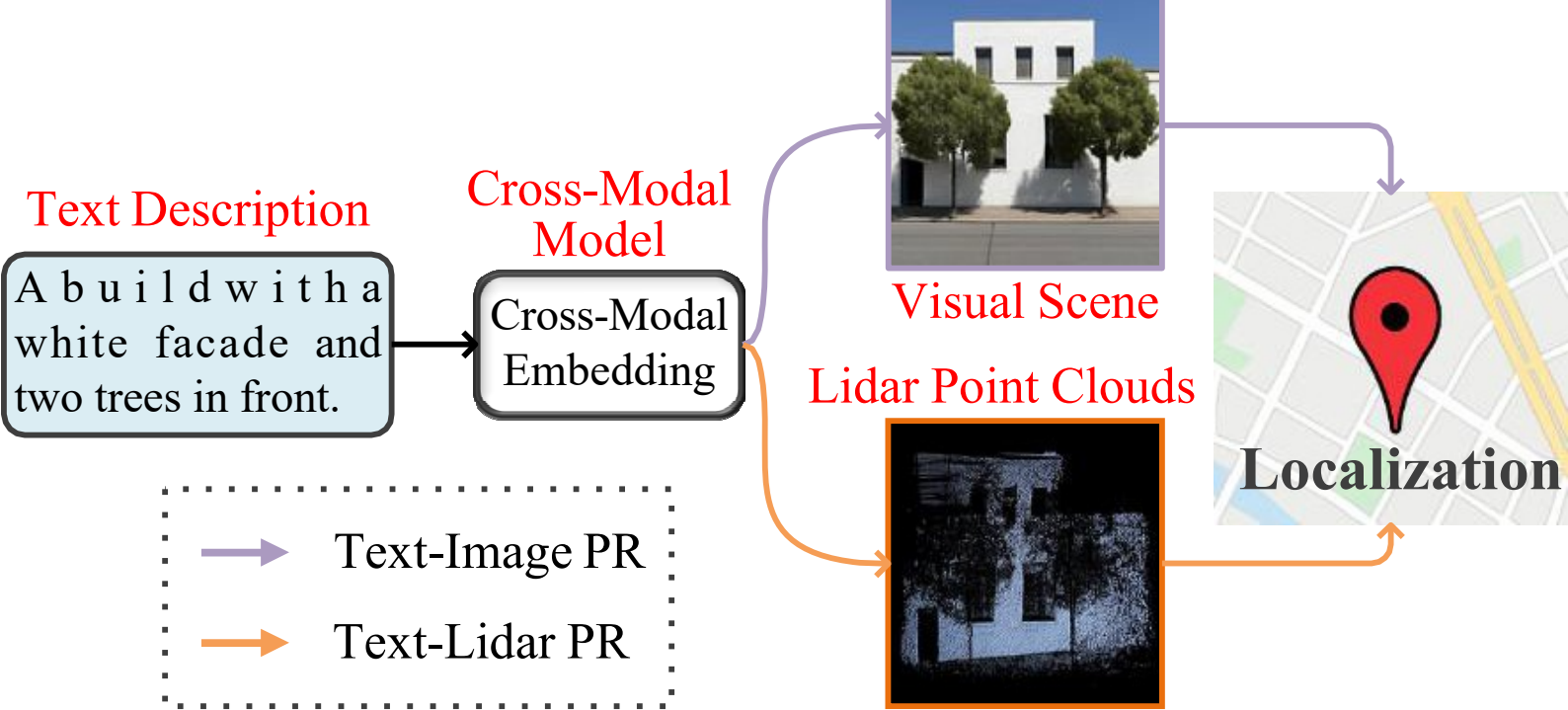

Figure 18: CMPR includes two main paradigms: Text-to-Image, which matches language to visual semantics via shared embedding spaces, and Text-to-LiDAR, which aligns textual descriptions with 3D geometric structures through direct localization or matching. While Text-to-Image methods excel in leveraging semantic richness, they are sensitive to appearance changes. In contrast, Text-to-LiDAR approaches offer stronger condition invariance due to the stability of point clouds but face greater challenges in cross-modal alignment due to sparse semantics.

**B. Text-image LPR Methods:** Text-image methods refer to cross-modal approaches that aim to bridge natural language and visual imagery, enabling mutual understanding, retrieval, and generation across modalities. These methods are foundational to vision-language pretraining, multimodal retrieval, image captioning, and text-to-image synthesis, and they are widely applied in information retrieval, robotics, content generation, and human-computer interaction.

CLIP [111] introduces a large-scale contrastive learning framework that jointly embeds images and texts into a shared semantic space using independent encoders. Although not originally designed for localization, CLIP laid the foundation for vision-language pretraining and has demonstrated strong performance in text-to-image retrieval and zero-shot transfer across modalities. Text4VPR [114] focuses on aligning natural language with multi-view images for large-scale place recognition. It employs frozen T5 embeddings, Sinkhorn optimal transport for matching, and cascaded cross-attention modules to achieve robust text-to-image registration. The approach demonstrates the feasibility of using free-form language as a query modality in visual place recognition.

**Remark 8:** *CMPR using natural language remains a nascent research area due to the significant semantic and representational gap between textual and spatial modalities. Two main approaches have emerged: (1) text-to-image localization, which maps language and visual data into a shared embedding space for scene-level matching, offering scalability but suffering from sensitivity to visual variations and ambiguity in language; and (2) text-to-Lidar localization, which directly regresses poses from text, enabling precise localization but facing challenges in data scarcity and weak geometric grounding. The inherent difficulty of cross-modal alignment continues to limit progress and widespread adoption in this field.*

# 4. Experiment Setup and Result Analysis

## *4.1. Datasets*

This survey summarizes place recognition datasets, including the datasets corresponding to VPR, LPR, and CMPR. It should be noted that these are only some of the frequently used ones. For a detailed summary, see Table 11.

Table 11: Summary of regularly used place recognition datasets: VPR datasets, LPR datasets, and CMPR datasets.

| Dataset | Modality | Number | Resolution | Application |
|---|---|---|---|---|
| Pittsburgh-30k [115] | Image | 30,000 | Varies | Google Street View |
| Pittsburgh-250k [115] | Image | 250,000 | Varies | Large-scale image retrieval benchmark |
| Mapillary val [116] | Image | 25,000 | 2-22 MP | Street-level images with pixel annotations |
| Mapillary test [116] | Image | 25,000+ | High-resolution | Urban scenes; Condition variation |
| St Lucia [117] | Image | Varies | Varies | Suburban driving scenes for VPR |
| Tokyo247 [118] | Image | 16,000 (DB) + 147 (query) | Varies | Day/night variation; City scenes |
| Nordland [119] | Image | 143,072 | Varies | Railway images across four seasons |
| Baidu Mall [120] | Image | 2,989 | High-res | Indoor shopping mall; Cross-device capture |
| 17 Places [121] | Image | Varies | Varies | Scene classification and retrieval |
| Nardo-Air R [122] | Image | Varies | $3464 \times 5202$ | Aerial nighttime RAW images |
| KITTI360Pose [22] | Text + LiDAR | 12,000+ | Velodyne HDL-64E | Text-driven 6-DoF pose localization |
| Street360Loc [114] | Text + Image | 10,000+ | Multi-view images | Cross-modal place recognition from language |
| Oxford [123] | LiDAR | Approximately 20,000 frames | Velodyne HDL-64E | Long-term LiDAR-based localization |
| U.S./R.A/B.D [86] | LiDAR | Varies | Velodyne HDL-64E | Multi-city LiDAR scans for localization |

### *4.1.1. VPR Datasets*

- Pittsburgh-30k/250k [115]: Collected from Google Street View, these datasets serve as standard benchmarks for large-scale urban image-based place recognition and retrieval. Pittsburgh-30k is a subset of the larger 250k version, both commonly used to evaluate retrieval performance in city-scale environments.

- Mapillary Vistas (val/test) [116]: A large-scale street-view dataset with high-resolution images and fine-grained pixel annotations across a wide geographic range. It is widely used for semantic understanding and robustness testing under varying environmental and lighting conditions.

- St Lucia [117]: Captured in a suburban area in Queensland, Australia. This dataset includes real driving sequences with dynamic scenes. It sup-ports evaluation of short-range visual localization and robustness to real-world traffic and environmental changes.

- Tokyo 24/7 [118]: Comprises image sequences captured at identical locations under daytime, twilight, and nighttime conditions. It is specifically designed to test the robustness of visual place recognition systems against severe illumination changes.

- Nordland [119]: Consists of over 140,000 images taken from a 728 km railway trip in Norway across four seasons. It serves as a benchmark for long-term visual localization and place recognition under extreme seasonal variation.

- Baidu Mall [120]: An indoor dataset collected in a shopping mall with training images taken using high-resolution cameras and test queries captured with mobile phones. It addresses cross-device, cross-time indoor localization challenges.

- 17 Places [121]: Contains scene-centric images from 17 indoor environments and is commonly used for scene classification and semantic-level image retrieval tasks.

- Nardo-Air R [122]: Provides aerial images at high resolution (up to 3464×5202 pixels), including nighttime raw data. It poses challenges for place recognition in aerial and remote sensing contexts, particularly under low-light conditions.

*4.1.2. LPR Datasets*

- Oxford RobotCar [123]: Collected over a year in Oxford, UK, with Velodyne HDL-64E and multiple sensors. It supports long-term LiDAR-based localization research under varying weather, seasons, and traffic conditions.

- U.S./R.A/B.D [86]: Composed of LiDAR scans from multiple U.S. cities,

this dataset is suitable for evaluating cross-city generalization in large-scale 3D localization.

#### 4.1.3. CMPR Datasets

- KITTI360Pose [22]: An extension of the KITTI-360 dataset that pairs natural language descriptions with LiDAR point clouds and annotated 6-DoF poses. It is the first dataset enabling direct language-to-3D localization and supports research in cross-modal geometric grounding and embodied navigation.

- Street360Loc [114]: Provides multi-view urban street-level images paired with natural language descriptions. It serves as a benchmark for cross-modal text-to-image place recognition and registration, supporting evaluation of language-grounded VPR across viewpoints.

### 4.2. Metric Evaluation

In place recognition tasks, two widely adopted evaluation metrics are the Precision-Recall (PR) Curve and Top@N accuracy, each capturing different aspects of system performance.

#### 4.2.1. PR Curve

Precision-Recall (PR) Curve is a fundamental metric that evaluates system performance across varying similarity thresholds. It captures the balance between precision, defined as the proportion of correctly retrieved locations among all retrieved candidates, and recall, the proportion of correctly retrieved locations among all relevant instances. Mathematically, these are expressed as:

$$Precision = \frac{TP}{TP + FP}, \quad Recall = \frac{TP}{TP + FN} \tag{1}$$

where $TP$ , $FP$ , and $FN$ denote true positives, false positives, and false negatives. By plotting precision against recall, the PR curve reflects how the system performs under different operating conditions. The area under this curve (AUC-PR) provides a scalar summary, particularly valuable for evaluating models in highly imbalanced datasets or when recall is critical.

*4.2.2. Recall@N*

In visual place recognition tasks, *Recall@N* is a widely used metric to evaluate the retrieval performance of a system. It measures the proportion of query images for which the correct matching image (i.e., ground truth) appears within the top $N$ retrieved candidates. Let $Q$ be the total number of query images, and for the $i^{th}$ query, let the top $N$ retrieved images be $\{r_{i1}, r_{i2}, ..., r_{in}\}$ denotes the ground truth image corresponding to the query. The *Recall@N* is then formally defined as:

$$Recall@N = \frac{1}{Q} \sum_{i=1}^{Q} \mathbb{1}(g_i \in \{r_{i1}, r_{i2}, \ldots, r_{iN}\}) \qquad (2)$$

where $\mathbb{1}(\cdot)$ is the indicator function that returns 1 if the condition is true, and 0 otherwise. A higher *Recall@N* value indicates that the system is more effective at retrieving the correct location within the top N candidates, making it a key metric for assessing the practical utility of image-based place recognition models.

## 5. Experimental Results

*5.1. Performance Evaluation of VPR Methods*

*5.1.1. Performance Evaluation of Sequence-based VPR Methods*

As shown in Figure 19, the sequence-based visual place recognition methods—SeqSLAM, FAB-MAP, DBoW, MRS-VPR, and OpenSeqSLAM 2.0—exhibit varying performance across datasets characterized by different environmental complexities. SeqSLAM and FAB-MAP demonstrate strong performance in stable environments with distinct features; however, they encounter difficulties in dynamic conditions, as evidenced by the Pittsburgh-30k and Mapillary tests. Notably, FAB-MAP, utilizing a probabilistic approach, slightly outperforms SeqSLAM in landmark-rich environments such as Oxford. DBoW, while efficient in feature-based environments, tends to underperform in noisy or occluded scenarios, as demonstrated by datasets such as Pittsburgh-250k. MRS-VPR excels in large-scale, appearance-variable datasets like Oxford and Nardo Air due to its multiresolution approach. OpenSeqSLAM 2.0 consistently outperforms its competitors, particularly in complex and dynamic datasets like Pittsburgh-250k and Nardo Air, owing to its advanced sequence matching techniques. It should be noted that some of the results are obtained by running the program. Since some parameter settings may not be consistent with the original paper, inconsistent results may occur.

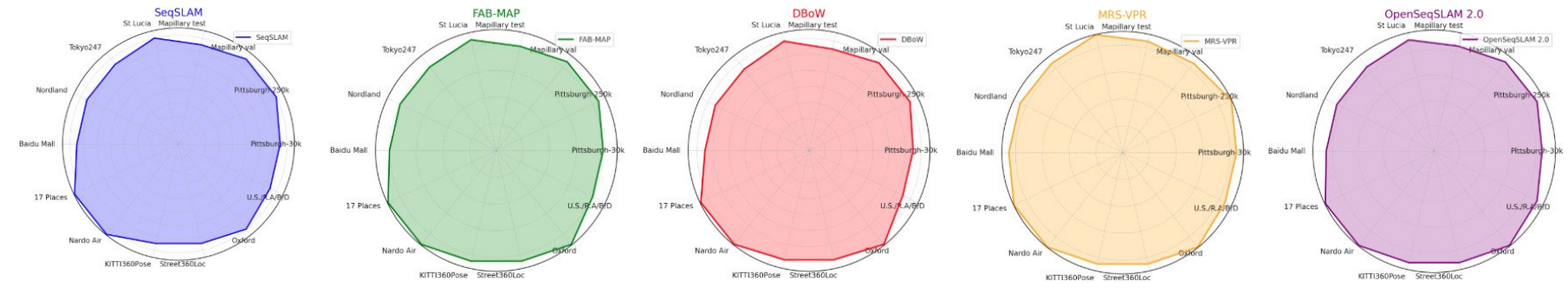

Figure 19: Performance evaluation of SeqSLAM, FAB-MAP, DBoW, MRS-VPR, and OpenSeqSLAM 2.0 across different datasets.

**Remark 9:** *In conclusion, OpenSeqSLAM 2.0 and MRS-VPR are more effective at managing large-scale and dynamic environments, whereas SeqSLAM, FAB-MAP, and DBoW are better suited for controlled environments with stable features. The selection of a method should be based on the complexity of the dataset and the variability of the environment.*

*5.1.2. Performance Evaluation of CNN-based VPR Methods*

The performance comparison across mainstream benchmark VPR datasets, shown in Figure 20 and Table 12, reveals distinct trends among recent SOTA CNN-based methods. Notably, EigenPlace and MixVPR consistently outperform other approaches, with EigenPlace achieving the highest *Recall@1* scores of 92.5% on Pitts30k and 92.4% on Tokyo 24/7, while MixVPR leads with scores of 94.6% on Pitts250k and 88.0% on MSLS-v. These results indicate that viewpoint-invariant representation learning (EigenPlace) and feature-mixing strategies (MixVPR) provide superior generalization across urban, cross-seasonal, and multi-perspective conditions. It should be noted that some of the results are obtained by running the program. Since some parameter settings may not be consistent with the original paper, inconsistent results may occur.

DELG, which integrates global and local features within a unified architec-ture, demonstrates exceptional performance on the Tokyo 24/7 dataset (*Recall@1* = 95.9%), underscoring its effectiveness in complex scenes with varying illumination. CosPlace and GSV-Cities, both trained on extensive street-view datasets, yield strong results on urban and geographically diverse datasets such as MSLS-c and SPED, with CosPlace achieving Recall@1 scores of 67.2% and 75.3%, respectively.

In contrast, traditional aggregation-based methods, such as NetVLAD and GeM, remain competitive on well-structured datasets (e.g., Pitts30k and Pitts250k). However, they tend to underperform on dynamic or seasonally variable datasets, such as Nordland and Tokyo 24/7. For instance, NetVLAD achieves only 32.6% *Recall@1* on Nordland, highlighting its limited robustness to significant appearance changes.

Interestingly, CRN exhibits a notable strength on MSLS-v (*Recall@1* = 82.2%), indicating that its attention-based mechanism improves performance in visually complex environments. However, its result on Nordland, reported as 44.8%, seems to be a typographical error and should be approached with caution.

**Remark 10:** *Overall, the results indicate a paradigm shift in VPR from traditional global descriptor methods to architectures that emphasize multi-scale feature aggregation, self-supervised training, and viewpoint invariance. The top-performing methods exhibit not only high retrieval accuracy in constrained urban environments but also robust resilience to challenging variations, including changes in weather, lighting, and perspective shifts.*

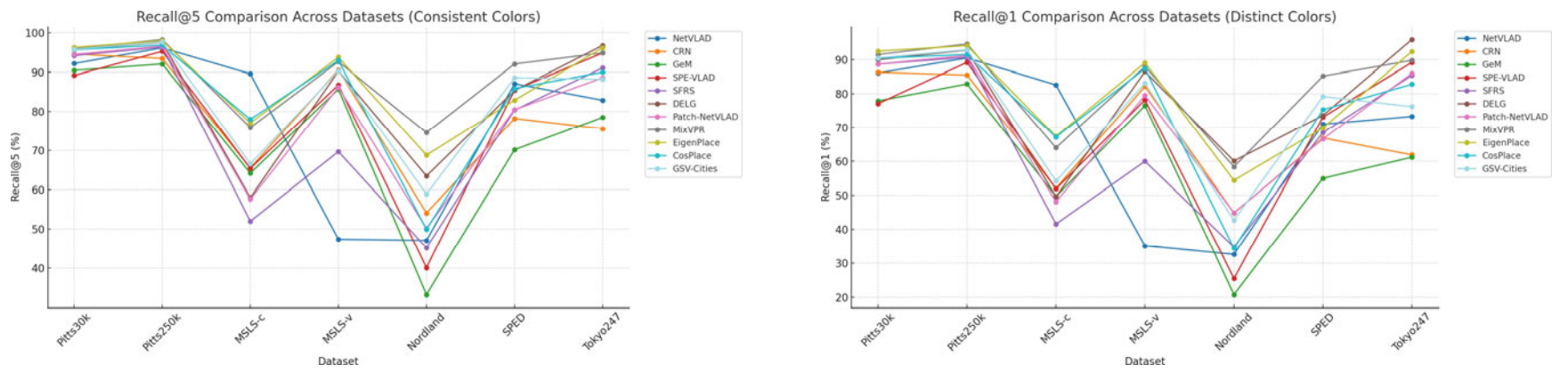

Figure 20: Recall@1 and Recall@5 of SOTA CNN-based VPR methods on mainstream benchmark datasets.

Table 12: Recall@1 and Recall@5 of SOTA CNN-based VPR methods on mainstream benchmark datasets. We did not include early simple CNN methods in the comparison, because their performance is too different from that of these. In addition, some methods do not have corresponding public codes, so we cannot evaluate them.

| Method | Pitts30k (%) | | Pitts250k (%) | | MSLS-c (%) | | MSLS-v (%) | | Nordland (%) | | SPED (%) | | Tokyo 24/7 (%) | |
|---|---|---|---|---|---|---|---|---|---|---|---|---|---|---|
| | R@1 | R@5 | R@1 | R@5 | R@1 | R@5 | R@1 | R@5 | R@1 | R@5 | R@1 | R@5 | R@1 | R@5 |
| NetVLAD [49] | 86.1 | 92.2 | 90.5 | 96.2 | 35.1 | 47.4 | 82.6 | 89.6 | 32.6 | 47.1 | 71.0 | 87.1 | 73.3 | 82.9 |
| CRN [50] | 86.3 | 94.6 | 85.5 | 93.5 | 52.1 | 65.5 | 82.2 | 90.6 | 44.8 | 54.1 | 66.9 | 78.2 | 61.9 | 75.6 |
| GeM [52] | 77.9 | 90.5 | 82.9 | 92.1 | 49.7 | 64.2 | 76.5 | 85.7 | 20.8 | 33.3 | 55.0 | 70.2 | 61.2 | 78.5 |
| SPE-VLAD [53] | 77.1 | 89.2 | 89.2 | 95.3 | 51.9 | 65.4 | 78.2 | 86.8 | 25.5 | 40.1 | 73.1 | 85.5 | 89.2 | 94.9 |
| SFRS [54] | 88.7 | 94.2 | 90.7 | 96.4 | 41.6 | 52.0 | 60.0 | 69.7 | 34.8 | 45.2 | 68.5 | 80.4 | 85.4 | 91.1 |
| DELG [55] | 90.0 | 95.7 | 92.8 | 97.6 | 49.6 | 58.0 | 86.5 | 90.3 | **60.1** | 63.5 | 73.6 | 85.4 | **95.9** | **96.8** |
| Patch-NetVLAD [57] | 88.7 | 94.5 | 91.2 | 96.6 | 48.1 | 57.6 | 79.5 | 86.2 | 44.9 | 50.2 | 66.6 | 80.5 | 86.0 | 88.6 |
| MixVPR [58] | 91.5 | **95.9** | **94.6** | **98.3** | 64.0 | 75.9 | 88.0 | 92.7 | 58.4 | **74.6** | **85.2** | **92.1** | 89.8 | 94.9 |
| EigenPlaces [59] | **92.5** | 96.3 | 94.1 | 98.0 | **67.4** | 77.1 | **89.1** | **93.8** | 54.4 | 68.8 | 69.9 | 82.9 | 92.4 | 96.2 |
| CosPlace [60] | 90.4 | 95.7 | 91.5 | 96.9 | 67.2 | **78.0** | 87.4 | 93.0 | 34.4 | 49.9 | 75.3 | 85.9 | 82.8 | 90.0 |
| GSV-Cities [61] | 90.5 | 95.8 | 92.9 | 97.7 | 54.2 | 66.6 | 83.1 | 90.3 | 42.7 | 58.9 | 79.2 | 88.6 | 76.2 | 88.2 |

### *5.1.3. Performance Evaluation of Transformer-based VPR Methods*

The comprehensive evaluation of SOTA Transformer-based VPR methods across benchmark datasets underscores the increasing maturity and specialization of the field, which is shown in Figure 21 and Table 13. Among these methods, Pair-VPR and SelaVPR++ consistently achieve top performance, with Pair-VPR attaining 100% *Recall*@1 on the Tokyo 24/7 dataset and maintaining strong results across urban, seasonal, and condition-varying datasets. This highlights the effectiveness of end-to-end, context-aware feature learning pipelines. SelaVPR++, which utilizes selective multi-scale attention, excels particularly in scenarios involving long-term variations, such as those presented in the Nordland dataset, demonstrating the advantages of hierarchical feature modeling. It should be noted that some of the results are obtained by running the program. Since some parameter settings may not be consistent with the original paper, inconsistent results may occur.

From a methodological perspective, recent Transformer-based VPR systems increasingly integrate the strengths of traditional global, local, and re-ranking approaches. Global methods, such as BoW and CricaVPR, provide robust and efficient retrieval through compact image-level descriptors. In contrast, local methods like SegVLAD and EffoVPR enhance structural sensitivity by incor-porating spatial and semantic granularity. Re-ranking pipelines, notably Pair-VPR, further refine coarse retrieval by utilizing context aggregation and feature interaction, significantly improving precision in challenging environments.

**Remark 11:** *The results highlight a distinct trend: top-performing methods no longer depend solely on global representations. Instead, they integrate multiple feature levels, often guided by visual semantics or spatial partitioning. This hybrid approach facilitates strong performance not only on large-scale urban datasets such as Pitts250k and MSLS but also under extreme visual transformations, as demonstrated on Nordland and Tokyo 24/7. Collectively, these advancements signify a paradigm shift toward multi-granularity, semantically enriched, and attention-guided VPR systems powered by Transformer architectures.*

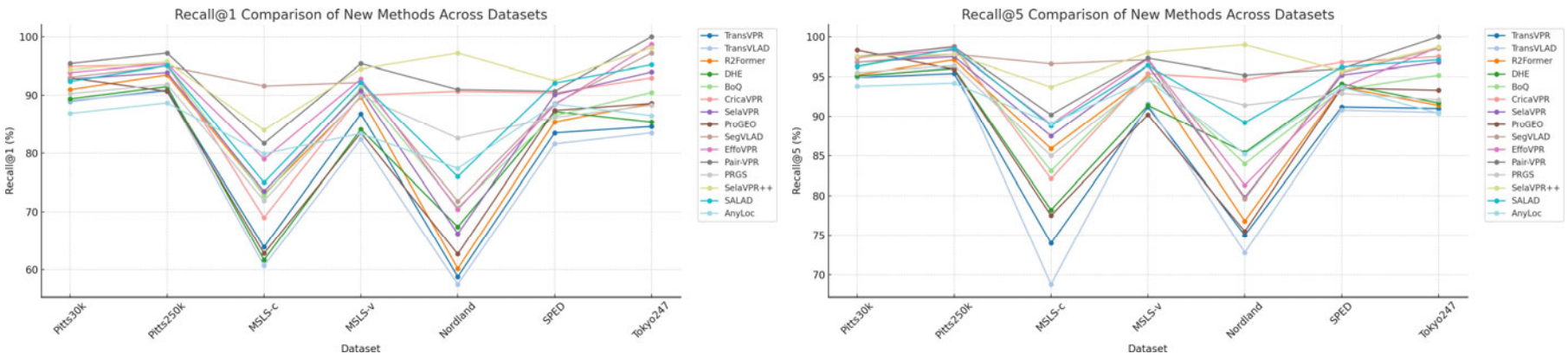

Figure 21: Recall@1 and Recall@5 of SOTA Transformer-based VPR methods on mainstream benchmark datasets.

*5.2. Performance Evaluation of LPR Methods*

*5.2.1. Performance Evaluation of Point-based LPR Methods*

A comparative analysis of eight representative point-based place recognition methods across five KITTI sequences (00, 02, 04, 06, and 08) shown in Figure 22 reveals distinct patterns in their robustness and generalization capabilities under varying scene conditions. It should be noted that some of the results are obtained by running the program. Since some parameter settings may not be consistent with the original paper, inconsistent results may occur.

Notably, TransLoc3D and MinkLoc3D consistently achieve the highest Recall@1 scores across all sequences. Their strong performance, particularly in complex environments such as KITTI04 and highly dynamic scenes like KITTI06, highlights the effectiveness of their multi-scale feature modeling and contextual spatial awareness. LPD-Net and DH3D also demonstrate stable results across most sequences, attributed to their ability to capture fine-grained local geometric structures, making them well-suited for structurally rich and repetitive urban scenes.

Table 13: Recall@1 and Recall@5 of SOTA Transformer-based VPR methods on mainstream benchmark datasets. Some methods do not have corresponding public codes, so we cannot evaluate them.

| Method | Pitts 30k | | Pitts250k | | MSLS-c | | MSLS-v | | Nordland | | SPED | | Tokyo 24/7 | |
|---|---|---|---|---|---|---|---|---|---|---|---|---|---|---|
| | R@1 | R@5 | R@1 | R@5 | R@1 | R@5 | R@1 | R@5 | R@1 | R@5 | R@1 | R@5 | R@1 | R@5 |
| TransVPR [62] | 89.0 | 94.9 | 90.9 | 95.4 | 63.9 | 74.0 | 86.8 | 91.2 | 58.8 | 75.0 | 83.5 | 91.2 | 84.6 | 91.0 |
| TransVLAD TransVLAD | 89.1 | 94.9 | 90.7 | 96.2 | 60.7 | 68.8 | 82.4 | 91.6 | 57.5 | 72.8 | 81.6 | 90.8 | 83.5 | 90.5 |
| R2Former [76] | 91. | 95.2 | 93.5 | 97.1 | 73.0 | 85.9 | 89.7 | 95.0 | 60.2 | 76.8 | 85.4 | 93.7 | 88.6 | 91.4 |
| DHE [66] | 89.4 | 95.1 | 91.5 | 96.0 | 61.7 | 78.2 | 84.1 | 91.4 | 67.4 | 85.4 | 87.2 | 94.1 | 85.4 | 91.7 |
| BoQ [67] | 92.4 | 96.2 | 95.0 | 98.5 | 72.8 | 83.1 | 91.2 | 95.3 | 70.7 | 84.0 | 86.5 | 93.4 | 90.5 | 95.2 |
| CricaVPR [68] | 94.9 | 97.3 | 95.3 | 98.8 | 69.0 | 82.1 | 90.0 | 95.4 | 90.7 | 94.6 | 90.4 | **96.8** | 93.0 | 97.5 |
| SelaVPR [69] | 92.8 | 96.8 | 93.9 | 97.5 | 73.5 | 87.5 | 90.8 | 96.4 | 66.2 | 79.8 | 90.1 | 95.2 | 94.0 | 96.8 |
| ProGEO [70] | 93.0 | **98.3** | 90.7 | 95.9 | 62.8 | 77.5 | 83.5 | 90.2 | 62.7 | 75.5 | 87.4 | 93.6 | 88.6 | 93.3 |
| SegVLAD [71] | 93.1 | 96.8 | 95.0 | 97.8 | - | - | - | - | 71.8 | 79.6 | 88.6 | 95.5 | 97.2 | 98.5 |
| EffoVPR [72] | 93.9 | 97.4 | 95.4 | 98.3 | 79.0 | 89.0 | 92.8 | 97.2 | 70.4 | 81.3 | 88.4 | 93.6 | 98.7 | 98.7 |
| Pair-VPR [20] | **95.4** | 97.5 | **97.2** | 98.7 | 81.7 | 90.2 | **95.4** | 97.3 | 91.0 | 95.2 | 90.7 | 96.0 | **100** | **100** |
| PRGS [73] | 90.3 | 95.5 | 91.9 | 96.4 | 71.9 | 85.0 | 89.9 | 94.6 | 82.6 | 91.4 | 86.2 | 92.9 | 88.3 | 92.1 |
| SelaVPR++ [74] | 94.4 | 97.5 | 96.5 | **99.0** | **84.0** | **93.7** | 94.5 | **98.0** | **97.2** | **99.0** | **92.9** | 96.7 | 98.1 | 98.7 |
| SALAD [19] | 92.4 | 96.3 | 95.1 | 98.5 | 75.0 | 88.8 | 92.2 | 96.4 | 76.0 | 89.2 | 92.1 | 96.2 | 95.2 | 97.1 |
| AnyLoc [75] | 86.9 | 93.8 | 88.7 | 94.2 | 79.9 | 89.1 | 83.4 | 94.6 | 77.4 | 85.2 | 88.5 | 93.7 | 86.5 | 90.4 |

In contrast, earlier methods such as PointNetVLAD show significantly lower performance, especially on sequences with structural variation, indicating limitations in local feature representation. PCAN, by incorporating attention mechanisms, offers moderate improvements but still lags behind more recent architectures in challenging scenarios.

HiTPR and SE(3)-Equivariant exhibit a strong balance of performance across all sequences, suggesting enhanced robustness to viewpoint changes and geometric transformations. Their integration of Transformer-based architectures and SE(3)-equivariance principles enables better generalization in cluttered and dynamic environments.

**Remark 12:** *Overall, the results emphasize the importance of incorporating*

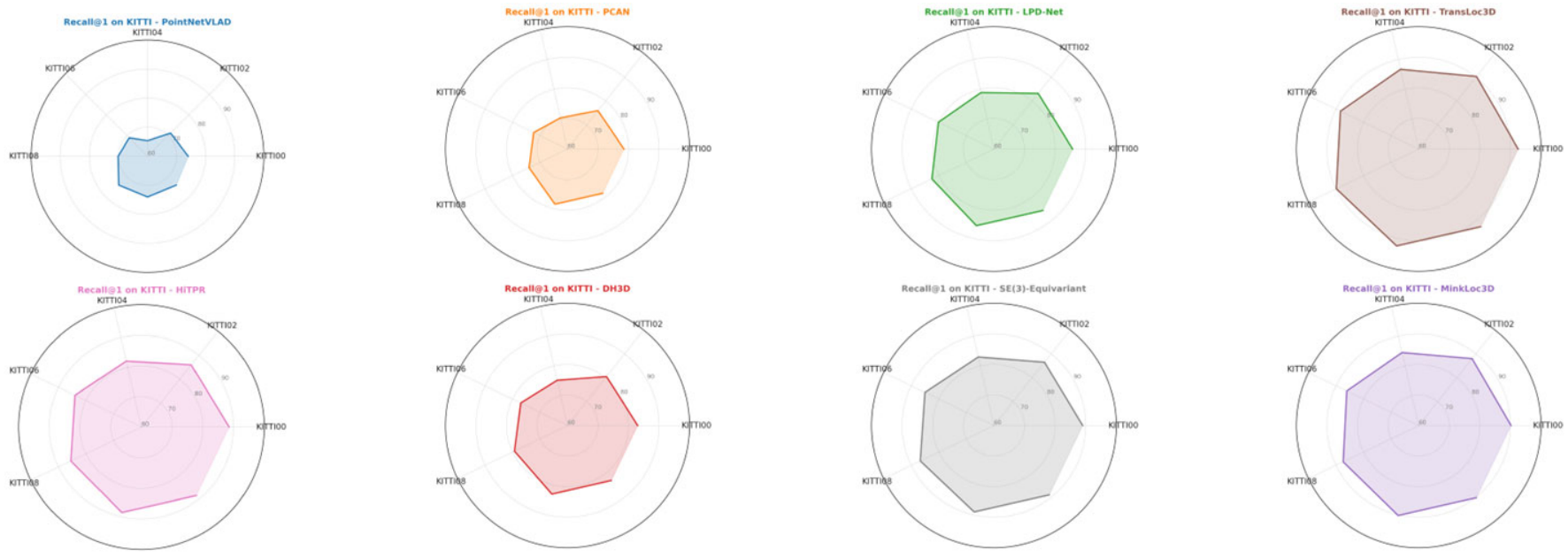

Figure 22: Performance evaluation of point-based LPR methods across different datasets.

*multi-scale representation learning, spatial reasoning, and transformation-aware encoding for robust place recognition in real-world point cloud data. Future research may further explore sparse computation and transformer-based hybrid architectures to optimize both accuracy and computational efficiency.*

*5.2.2. Performance Evaluation of Transformer-based LPR Methods*

Based on the comparative radar charts (see Figure 22), several key observations can be made regarding the performance of SOTA Transformer-based LPR methods across benchmark datasets (Oxford, U.S., R.A., B.D., and KITTI). It should be noted that some of the results are obtained by running the program. Since some parameter settings may not be consistent with the original paper, inconsistent results may occur.

SVT-Net and SALSA exhibit consistently superior Recall@1 performance across all datasets, particularly excelling in large-scale and diverse environments such as Oxford and U.S., where their accuracy approaches. These results underscore the effectiveness of combining sparse voxel representations with lightweight Transformer modules (SVT-Net) and the utility of radial-window attention within a scalable backbone (SALSA) for learning global context from sparse 3D data.

TransLoc3D, OverlapTransformer, and OPAL also demonstrate strong and stable performance, especially in dynamic and structurally complex environ-ments like B.D. and KITTI. Their architectures benefit from adaptive receptive fields, yaw-invariant attention, and multimodal fusion with semantic priors (e.g., OpenStreetMap data in OPAL), contributing to enhanced robustness under viewpoint variations and occlusions.

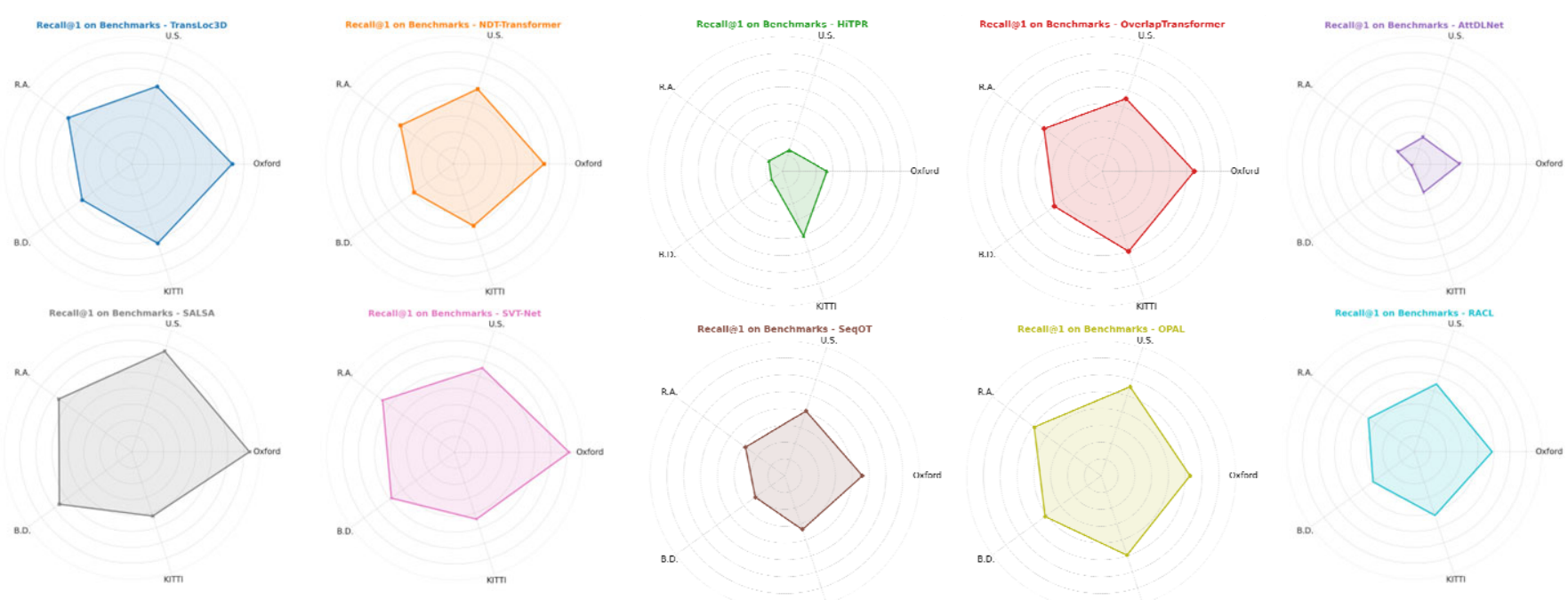

Figure 23: Performance evaluation of Transformer-based LPR methods across different datasets.

In contrast, AttDLNet and HiTPR, while offering computational efficiency and lightweight design, show comparatively lower accuracy, particularly in datasets with repetitive or ambiguous structures. This indicates that although these models are suitable for real-time applications, they may require additional enhancements to achieve state-of-the-art precision in challenging scenes.

SeqOT and RACL stand out in temporally and spatially dynamic environments, highlighting the advantages of incorporating sequential cues and continual learning mechanisms. These approaches are particularly relevant for applications such as autonomous navigation and long-term localization, where robustness to changing conditions is essential.

**Remark 13:** *Overall, the results highlight that methods capable of modeling global context, geometric structure, and semantic or temporal information tend to achieve higher performance across varied real-world conditions. Future research may benefit from further integration of efficient Transformer designs, sparsity-aware computation, and cross-modal reasoning to enhance the scalability and generalization of place recognition systems.*

5.3. Performance Evaluation of CMPR Methods

The comparative analysis of recall performance across validation and test sets, shown in Figure 24 and Table 14, reveals distinct trends between text-image and text-Lidar cross-modal place recognition methods. Text-image approaches, particularly Text4VPR, demonstrate superior performance and robustness across all k-values, consistently achieving the highest recall and maintaining strong generalization from validation to test environments. This highlights the effectiveness of vision-language pretraining and panoramic image matching for large-scale, semantically rich scenes. In contrast, text-LiDAR methods such as RET and Des4Pos exhibit competitive performance at higher k-values but show reduced accuracy at k=1, indicating challenges in precise point-level localization. Among them, Des4Pos benefits from Transformer-based modality fusion, outperforming RET in both stability and accuracy.

**Remark 14:** *Overall, text-image methods tend to generalize better in diverse urban environments, while text-LiDAR approaches are more geometry-sensitive and may be preferable in applications requiring high spatial fidelity. These findings suggest that future work should explore hybrid architectures that integrate image and LiDAR modalities under unified textual supervision to enhance both precision and robustness in cross-modal place recognition.*

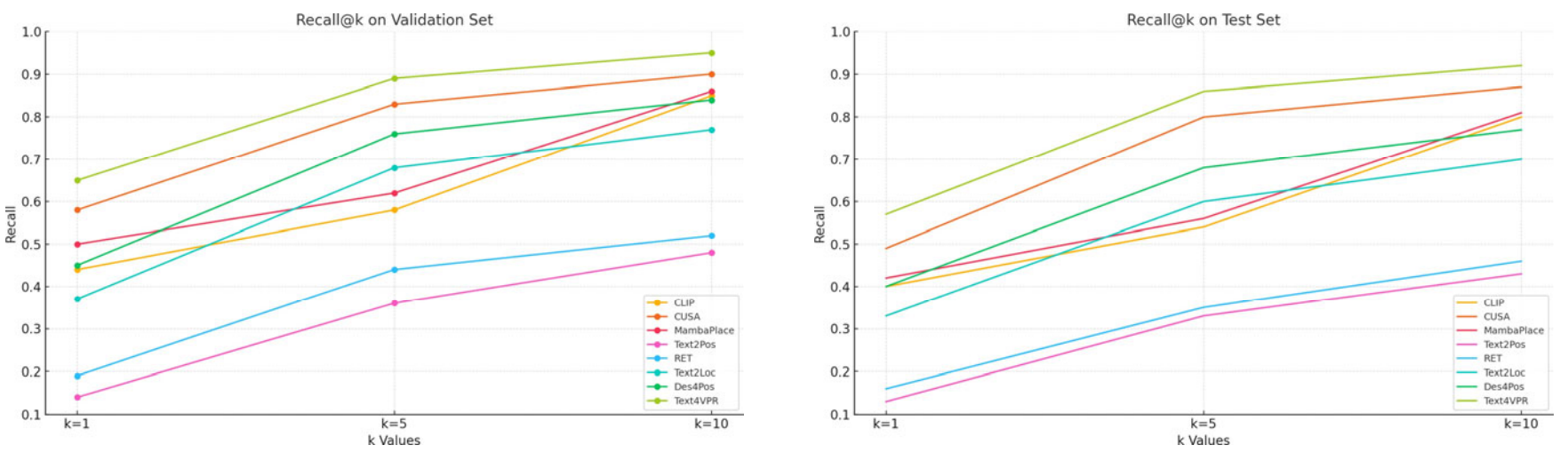

Figure 24: Recall@1, 5, 10 of SOTA CMPR methods on benchmark datasets.

Table 14: Comprehensive comparison of existing SOTA CMPR methods on public datasets.

| **Method** | **2*Dataset** | **Validation Set** | | | **Test Set** | | |
|---|---|---|---|---|---|---|---|
| | | **k=1** | **k=5** | **k=10** | **k=1** | **k=5** | **k=10** |
| CLIP [111] | Street360Loc | 0.44/0.45/0.48 | 0.58/0.59/0.60 | 0.85/0.87/0.87 | 0.40/0.41/0.45 | 0.54/0.55/0.56 | 0.80/0.81/0.82 |
| CUSA [124] | Street360Loc | 0.58/0.60/0.6 | 0.83/0.85/0.85 | 0.90/0.90/0.92 | 0.49/0.52/0.5 | 0.80/0.82/0.83 | 0.87/0.88/0.90 |
| MambaPlace [112] | KITTI360Pose | 0.50/0.50/0.56 | 0.62/0.63/0.6 | 0.86/0.87/0.88 | 0.42/0.44/0.46 | 0.56/0.58/0.58 | 0.81/0.82/0.83 |
| Text2Pos [22] | KITTI360Pose | 0.14/0.25/0.31 | 0.36/0.55/0.61 | 0.48/0.68/0.74 | 0.13/0.20/0.30 | 0.33/0.42/0.49 | 0.43/0.61/0.65 |
| RET [23] | KITTI360Pose | 0.19/0.30/0.37 | 0.44/0.62/0.67 | 0.52/0.72/0.78 | 0.16/0.25/0.29 | 0.35/0.51/0.56 | 0.46/0.65/0.71 |
| Text2Loc [21] | KITTI360Pose | 0.37/0.57/0.63 | 0.68/0.85/0.87 | 0.77/0.91/0.93 | 0.33/0.48/0.52 | 0.60/0.75/0.78 | 0.70/0.84/0.86 |
| Des4Pos [113] | KITTI360Pose | 0.45/0.63/0.69 | 0.76/0.89/0.92 | 0.84/0.94/0.96 | 0.40/0.54/0.57 | 0.68/0.80/0.82 | 0.77/0.87/0.89 |
| Text4VPR [114] | Street360Loc | **0.65/0.67/0.74** | **0.89/0.88/0.91** | **0.95/0.96/0.96** | **0.57/0.60/0.66** | **0.86/0.87/0.89** | **0.92/0.93/0.94** |

## 6. Challenges and Solutions

Place recognition is a fundamental component of autonomous navigation and SLAM, enabling a robot to recognize previously visited locations. Depending on the sensing modality, place recognition can be broadly categorized into VPR, LPR, and CMPR. Each modality offers unique advantages but also introduces specific challenges. This section presents an analysis of these challenges and corresponding solutions across the three categories. Each modality of place recognition exhibits unique advantages and corresponding limitations, as summarized in Table 15.

Table 15: Summary of challenges produced from different modalities.

| **Method** | **Advantages** | **Key Challenges** |
|---|---|---|
| VPR | Rich semantic information, low cost | Appearance change, viewpoint variation, dynamic occlusion |
| LPR | Lighting invariance, geometric fidelity | Sparsity, occlusion, computational burden |
| CMPR | Complementary information, robustness | Modality gap, feature alignment, data scarcity |

### *6.1. Challenges*

#### *6.1.1. Challenges for VPR*

VPR relies primarily on images to retrieve or recognize previously visited locations. Despite the significant progress enabled by deep learning-based feature extraction and matching methods, several critical challenges persist:

- Appearance Variation: Images are highly sensitive to changes in illumination, weather, seasonal dynamics, and time of day. These variations can lead to significant discrepancies in the appearance of the same place, which undermines the robustness of image-based matching. Although data augmentation and contrastive learning strategies have been proposed to improve invariance, extreme conditions, such as nighttime versus daytime, remain problematic.

- Viewpoint Variation: Due to the mobility of robots or vehicles, the same location may be observed from drastically different viewpoints. Such variation results in limited overlap of visual content, posing a significant challenge for both local and global feature matching.

- Dynamic Objects and Occlusions: Urban and human-populated environments introduce a large number of dynamic entities such as vehicles and pedestrians. These non-static objects may obscure static scene elements or introduce spurious features, leading to false positives or retrieval failures.

*6.1.2. Challenges for LPR*

Lidar-based place recognition utilizes 3D point cloud data, which offers robustness against appearance-related variations (e.g., lighting). However, it faces its own set of modality-specific challenges.

- Sparsity and Non-Uniform Sampling: Lidar point clouds are often sparse and vary in density depending on the range and sensor characteristics. Inconsistent sampling between scans can degrade the stability of geometric descriptors and matching performance.

- Occlusion and Structural Changes: Physical obstructions, such as vehicles or vegetation, can result in incomplete or occluded point clouds. Additionally, long-term environmental changes (e.g., construction or terrain modifications) can significantly affect the geometric layout, making long-term place recognition difficult.

- Computational Efficiency and Registration Sensitivity: Processing high-dimensional point clouds is computationally intensive, especially in large-scale retrieval scenarios.

### 6.1.3. Challenges for CMPR

CMPR aims to establish correspondences between different sensing modalities, such as images and point clouds. These approaches are promising for enhancing robustness but introduce a distinct set of challenges.

- Modality Gap: Developing effective architectures for modality fusion or alignment (e.g., through attention mechanisms or contrastive learning) requires careful design and significant computational resources. Misalignment during training can lead to poor generalization.

- Data Pairing and Supervision Scarcity: Cross-modal training often relies on accurately paired data (e.g., RGB images/point clouds aligned with text descriptions), which are difficult to obtain in large volumes. The lack of well-annotated cross-modal datasets limits the effectiveness of supervised learning approaches. Unsupervised or weakly supervised techniques are emerging but remain unstable in practice.

## 6.2. Possible Solutions/Future Directions

PR remains a core yet challenging problem in autonomous navigation, with distinct limitations across visual, Lidar, and cross-modal paradigms. Based on the analysis of all methods, the following research directions that may solve the problem can be derived:

- VPR is vulnerable to appearance and viewpoint changes, prompting advances in contrastive learning, self-supervised representation learning, and attention-based filtering of dynamic content.

- LPR, while robust to illumination, faces challenges from point cloud sparsity, occlusion, and computational overhead, motivating the use of learned geometric descriptors and efficient retrieval schemes.

- CMPR aims to bridge heterogeneous modalities such as vision and geometry but is hindered by modality gaps, feature alignment difficulty,

and limited supervision. Recent trends emphasize unified representation spaces, attention-driven fusion, and domain-adaptive training.

**Remark 15:** *Future research is expected to focus on multi-modal, self-supervised, and transformer-based frameworks to enable robust and scalable place recognition in diverse, long-term scenarios.*

## 7. Conclusions

Focusing on the advancement of robust place recognition across heterogeneous sensing modalities, this work comprehensively reviews the representative methodologies and core challenges in VPR, LPR, and CMPR. These approaches are systematically categorized based on their primary data modalities and algorithmic innovations. Through comparative analysis, this review reveals that VPR methods exhibit strong semantic perception but remain sensitive to variations in appearance and viewpoint, while LPR methods demonstrate structural robustness but suffer from issues related to sparsity and occlusion. CMPR methods attempt to unify the strengths of both, but face difficulties in modality alignment and representation consistency. In addition, this work contrasts recent representative models in terms of feature invariance, retrieval efficiency, and adaptability under varying environmental conditions. Experimental insights from existing benchmarks suggest that hybrid and transformer-based architectures show promising generalization across modalities and conditions. Finally, this review outlines future directions, including cross-modal contrastive learn- ing, self-supervised adaptation, and the design of unified architectures to further enhance the scalability, robustness, and generalization ability of place recognition systems in complex and dynamic environments. This work aims to provide a clear and comparative perspective to guide future research in the field.

## CRediT authorship contribution statement

**Zhenyu Li:** Writing – review & editing, Writing – original draft, Methodology, Investigation, Formal analysis, Data curation, Conceptualization. **Tianyi**

**Shang:** Writing – original draft, Conceptualization. **Pengjie Xu:** Supervision, Funding acquisition, Conceptualization. **Zhaojun Deng:** Visualization, Validation, Software, Resources.

## Declaration of Generative AI and AI-assisted technologies in the writing process

During the preparation of this work, the authors used ChatGPT in order to improve language and readability. After using this tool/service, the authors reviewed and edited the content as needed and take full responsibility for the content of the publication.

## Declaration of competing interest

The authors declare that they have no known competing financial interests or personal relationships that could have appeared to influence the work reported in this paper.

## Acknowledgment

This work was funded by the Shandong Provincial Natural Science Foundation (ZR2024QF284), the Chinese Society of Construction Machinery Young Talent Lifting Project (CCMS-YESS2023001), the Opening Foundation of Key Laboratory of Intelligent Robot (HBIR202301), the Open Project of Fujian Key Laboratory of Spatial Information Perception and Intelligent Processing (FKL-SIPIP1027).